\documentclass[journal,twoside,web]{ieeecolor}
\usepackage{epstopdf}
\usepackage{jsen}
\usepackage{cite}
\usepackage{amsmath,amssymb,amsfonts}
\usepackage{graphicx}
\usepackage{textcomp}
\usepackage{wrapfig}
\usepackage[backref]{hyperref}
\usepackage{threeparttable}
\usepackage[table,xcdraw]{xcolor}
\usepackage[normalem]{ulem}
\useunder{\uline}{\ul}{}
\usepackage{makecell}
\usepackage{threeparttable}
\usepackage[ruled,linesnumbered]{algorithm2e}

\def\BibTeX{{\rm B\kern-.05em{\sc i\kern-.025em b}\kern-.08em
    T\kern-.1667em\lower.7ex\hbox{E}\kern-.125emX}}
\definecolor{abstractbg}{rgb}{0.89804,0.94510,0.83137}
\begin{document}
\title{Clear Memory-Augmented Auto-Encoder for Surface Defect Detection}
\author{Wei Luo$^\ast$, \IEEEmembership{Student Member, IEEE}, Tongzhi Niu$^\ast$, \IEEEmembership{Graduate Student Member, IEEE}, Lixin Tang$^\dag$, Wenyong Yu and Bin Li, \IEEEmembership{Member, IEEE}
\thanks{Manuscript received XX XX, 20XX; revised XX XX, 20XX. This study was financially supported by the National Natural Science Foundation of China under Grant 51775214. (Corresponding author: Lixin Tang.)}
\thanks{Wei Luo, Tongzhi Niu, Lixin Tang, Wenyong Yu and Bin Li are with the State Key Laboratory of Digital
Manufacturing Equipment and Technology, School of Mechanical Science and Engineering, Huazhong University of Science and Technology, Wuhan 430074, China (e-mail: u201910709@hust.edu.cn, tzniu@hust.edu.cn, lixintang@hust.edu.cn, ywy@hust.edu.cn, libin999@hust.edu.cn).}
\thanks{$\ast$Equal contribution.}
\thanks{$\dag$Corresponding author.}}

\IEEEtitleabstractindextext{
\fcolorbox{abstractbg}{abstractbg}{
\begin{minipage}{\textwidth}
\begin{wrapfigure}[12]{r}{3in}
\centering
\includegraphics[width=2.8in]{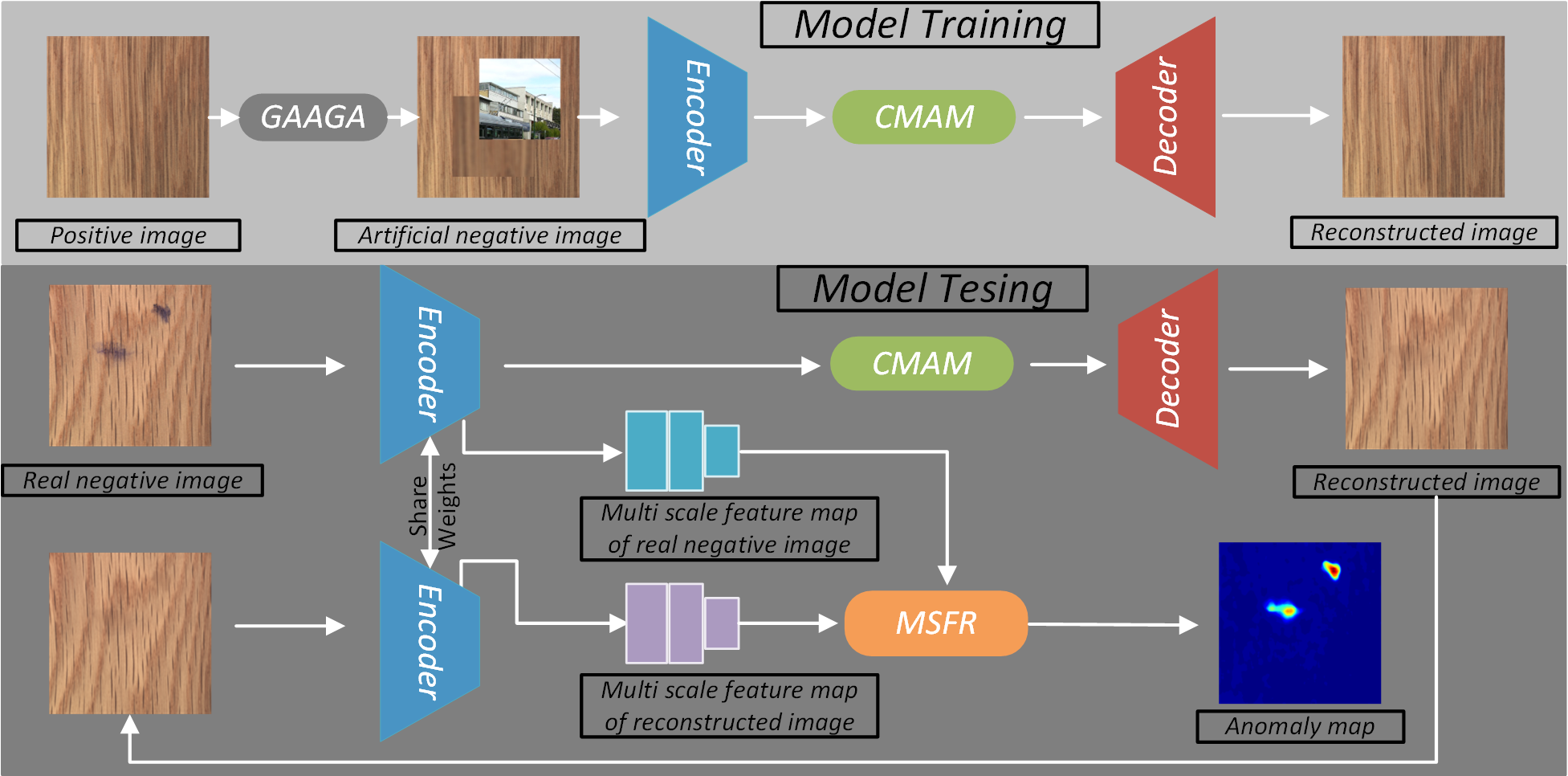}
\end{wrapfigure}
\begin{abstract}
In surface defect detection, due to the extreme imbalance in the number of positive and negative samples, positive-samples-based anomaly detection methods have received more and more attention. Specifically, reconstruction-based methods are the most popular. However, existing methods are either difficult to repair abnormal foregrounds or reconstruct clear backgrounds. Therefore, we propose a clear memory-augmented auto-encoder (CMA-AE). At first, we propose a novel clear memory-augmented module (CMAM), which combines the encoding and memory-encoding in a way of forgetting and inputting, thereby repairing abnormal foregrounds and preserving clear backgrounds. Secondly, a general artificial anomaly generation algorithm (GAAGA) is proposed to simulate anomalies that are as realistic and feature-rich as possible. At last, we propose a novel multi scale feature residual detection method (MSFR) for defect segmentation, which makes the defect location more accurate. Extensive comparison experiments demonstrate that CMA-AE achieves state-of-the-art detection accuracy and shows great potential in industrial applications.
\end{abstract}

\end{minipage}}}

\maketitle

\section{Introduction}
\label{sec:introduction}
\IEEEPARstart{I}{n} the industrial field, due to the complexity of the manufacturing process, surface defects are common in some industrial products such as fabric \cite{fabric} and steel \cite{steel}. These defects not only lead to poor user experience but also may cause industrial accidents. For example, surface defects of steel may reduce the Contact Fatigue Strength of the material. Therefore, defects inspection is an important method to achieve quality management.
Over the past decades, various surface defect detection methods have been proposed. These methods can be broadly divided into two categories, including traditional methods and deep-learning-based methods. Traditional methods mainly extract features manually and set thresholds to detect defects. Aiger and Talbot \cite{PHOT} proposed Phase-Only Transform (PHOT) to detect irregular defects in the background. Xie and Mirmehdi \cite{TEXTEMS} presented a Gaussian mixed model to detect and localize defects. Low-pass filtering with curvature analysis (LCA) was proposed by Tsai and Huang \cite{LCA} to remove periodic textural patterns. However, the feature extraction ability of the above methods is limited and the robustness is poor. As a data-driven method, deep learning can automatically extract features by training a large amount of data, with strong feature extraction ability and good generalization.

Most deep-learning-based methods are supervised learning approaches. PGA-Net \cite{PGANet} proposes a pyramid feature fusion and global context attention network for pixel-wise detection. Tabernik \textit{et al}. \cite{Seg-based} presented a segmentation-based deep learning architecture for surface defect detection. However, the above methods require a large number of defective samples and corresponding labels, which is difficult to obtain in the industrial field. Firstly, the number of defect-free samples is much larger than that of defective samples, and unknown defect types may occur in the process of production. Therefore, data collection becomes the biggest challenge in defect detection tasks. Secondly, sample labeling requires experienced engineers to spend a lot of time and effort, which is time-consuming and high-labor cost. These factors above limit the application of supervised learning in the industrial fields. 

In contrast, unsupervised learning methods show great potential as it only requires unlabeled normal samples for training. In recent years, unsupervised anomaly detection has received increasing attention. Many anomaly detection methods are applied to defect detection and made great progress. RIAD \cite{RIAD} treats anomaly detection as a reconstruction-by-inpainting problem and achieved excellent results in image anomaly detection. Zhang \textit{et al}. \cite{zhang2022deep} proposed a deep adversarial anomaly detection method with task-specific features for anomaly detection. Cho \textit{et al}. \cite{cho2022unsupervised} proposed a normal features distribution model in an unsupervised manner for video anomaly detection. Ribeiro \textit{et al}. \cite{ribeiro2018study} proposed a convolutional auto-encoder architecture to learn normal behavior and use the model for anomaly detection. Zhou \textit{et al}. \cite{zhou2022discovering} proposed a patch-level autoencoder combined with a context-enhanced autoregressive network for anomaly detection. Chen \textit{et al}. \cite{chen2021nm} proposed an end-to-end pipeline named NM-GAN, which assembles an encoder-decoder reconstruction network and a CNN-based discriminative network. Yang \textit{et al}. \cite{yang2022learning} mainly proposed deep features corresponding to anomaly detection and segmentation.

\begin{figure}
    \centering
    \includegraphics{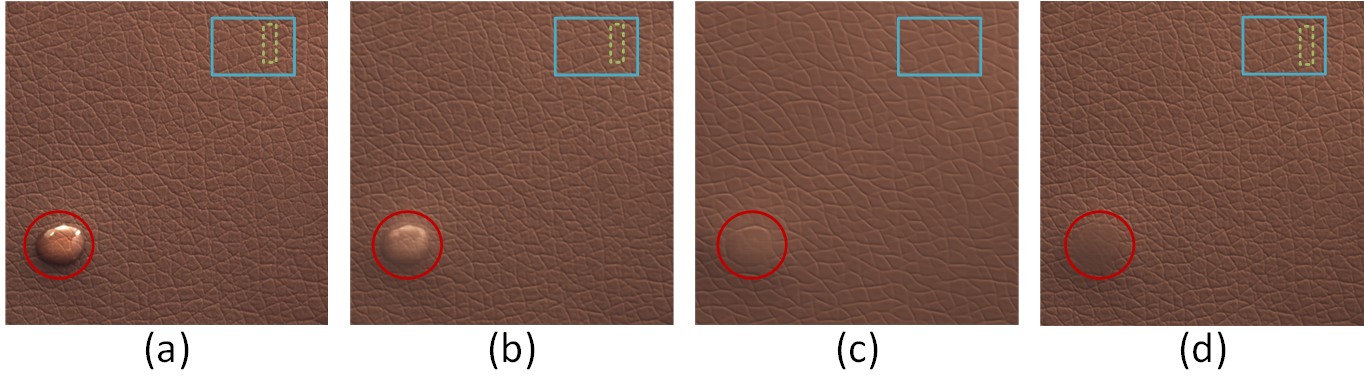}
    \caption{Reconstruction results of different methods. (a) is the original image. (b), (c), and (d) are the reconstructed images using AE \cite{AE}, MemAE \cite{MemAE}, and our proposed method CMA-AE, respectively. The red circles and green dotted boxes indicate defects and textures, respectively.}
    \label{cam-ae-example-fig}
\end{figure}

For anomaly detection, trained on the normal samples, the model is expected to produce larger reconstruction/generate errors for the anomalies than that for the normal ones. Therefore, the two most important capabilities of the model are the ability to reconstruct normal backgrounds and the ability to repair abnormal foregrounds. Because of the under-designed representation of the latent space, common AE \cite{AE} has the capability of the former, but not the latter. MemAE \cite{MemAE} proposes an improved AE that repairs the abnormal foregrounds by editing latent codes with a memory-augment module. The proposal of MemAE greatly enhances the repair ability of abnormal foreground but weakens the ability to reconstruct normal background at the same time, as shown in Fig. \ref{cam-ae-example-fig}. Specifically, MemAE uses the encoding from the encoder as a query to retrieve the most relevant memory items for decoding, which completely replaces the input features, resulting in a loss of detail. Therefore, we propose a novel clear memory-augmented module (CMAM), which combines encoding and memory-encoding in a way of inputting and forgetting, thereby repairing abnormal foregrounds and preserving clear backgrounds.

There are two improvements in the CMAM. At first, inspired by LSTM \cite{LSTM}, the way of inputting and forgetting is designed to improve the memory mechanism. We try to erase anomalous foregrounds in coding through a forget gate, and then use the memory-encoded information to inpainting the erased features through an input gate, resulting in a clear reconstruction map. Secondly, to learn how to forget and input, we proposed a two-stage training strategy. In the first stage, training with normal samples, the memory content is updated together with the encoder and decoder. In the second stage, the memory content is no longer updated, and the input and forgetting abilities are learned by repairing artificial anomalies in the artificial abnormal samples. 

Recently, artificial anomalies have been widely used to enhance the models. AFEAN \cite{AFEAN} generates artificial anomalies by combining defect-free images and random masks. Lv \textit{et al}. \cite{lv2020novel} proposed to use redundant features in natural images to simulate defects. Cutpaste \cite{cutpaste} cuts a small rectangular area from a normal training image and pastes it back to an image at a random location. But as all as we know, existing artificial anomalies are designed based on human experience and can only simulate limited real anomalies. Therefore, we propose a general artificial anomaly generation algorithm (GAAGA) to simulate anomalies that are as realistic and feature-rich as possible. We first assume that the features of natural images are redundant enough to simulate almost all anomalous features. And the blur of normal backgrounds can be regarded as degenerate anomalies. Inspired by Cutpaste, natural images and the blurring images of normal backgrounds as image patches are both pasted at a random location of a large image. 

Finally, in anomaly detection, the pixel gap between the original image and the reconstructed image is still used for defect segmentation in anomaly detection, which will cause a lot of noise and lead to the occurrence of false detection. Individual pixel has no semantics, normal and abnormal are context-dependent semantic descriptions. Whether it is filters of traditional methods or convolutional neural networks, contextual relevance is considered to be the key to image processing. Therefore, replacing pixel gaps with feature gaps is a more feasible approach. Since the size of the anomaly is ambiguous, we proposed a novel multi scale feature residual (MSFR) detection method for defect segmentation. 

There are two advantages in MSFR. Firstly, the residual between the original feature map and the reconstructed feature map guarantees the correlation between pixels, since an element in the feature map corresponds to pixels in one region of the original image. Secondly, different feature maps have different receptive fields, multi scale feature map residuals are used to obtain multi-scale information.
To sum up, the main contributions of our work are as follows:
\begin{itemize}
    \item We propose a novel clear memory-argument module (CMAM), which solves the problem of poor normal background reconstruction in MemAE \cite{MemAE}. A two-stage training strategy is adopt to improve the ability to reconstruct normal background and the ability to repair abnormal foreground respectively.
    \item We propose a novel multi scale feature residual method for defect segmentation (MSFR), which effectively solves the noise problem caused by using the pixel gap between the original image and the reconstructed image for defect segmentation, and makes the defect location more accurate.
    \item We also propose a novel artificial anomaly generation algorithm to simulate various possible real defects in industry.
\end{itemize}

The rest of this article is organized as follows. Section \MakeUppercase{\romannumeral2} introduces the related works of anomaly detection. In Section \MakeUppercase{\romannumeral3}, the proposed CMA-AE and MSFR are elaborated and discussed in detail. Section \MakeUppercase{\romannumeral4} designs extensive experiments to demonstrate the performance of CMA-AE and MSFR. Finally, Section \MakeUppercase{\romannumeral5} summarizes the paper.

\begin{figure*}[!t]
    \centering
    \includegraphics[width=188mm]{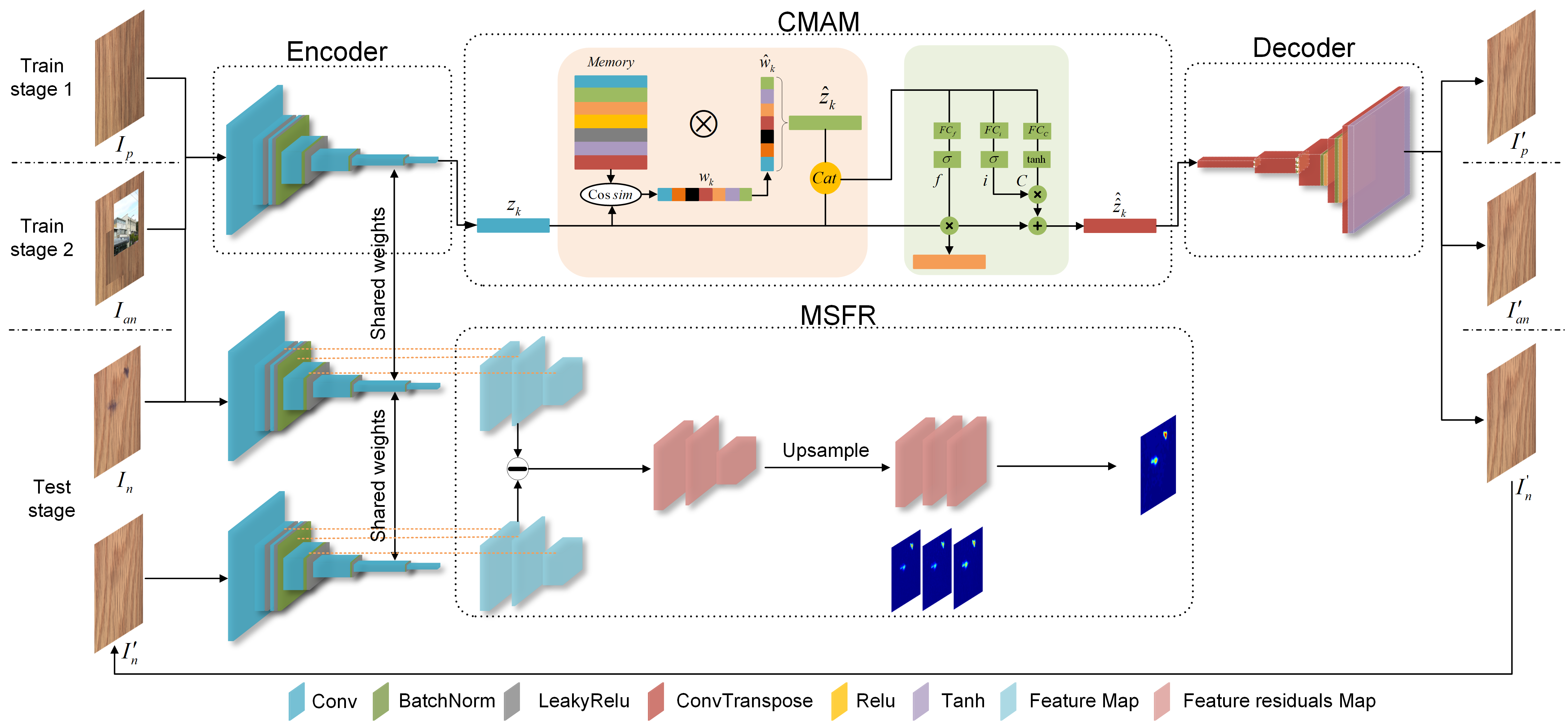}
    \caption{Overall architecture of the proposed CMA-AE in the train and test stages. CMA-AE consists of a CMAM, a MSFR, an encoder and a decoder. During training, positive samples and artificial negative samples are propagated forward in two stages. During testing, the abnormal area of negative samples is obtained by MSFR.}
    \label{networks-fig}
\end{figure*}
\section{Relate Works}
In recent years, the success of deep learning model training is generally determined by the number of representative training samples and the quality of the annotation. Therefore, anomaly detection based on positive samples without labels have received more and more attention. Many reconstruction-based anomaly detection methods have been proposed, most of which can be classified into two categories, including Auto-Encoder (AE) \cite{AE}, Generate Adversarial Nets (GANs) \cite{GANs} and their variants.

AE-based models can learn encoded features in the latent feature domain by training normal samples, and then reconstruct the background images from them. During testing, the output of anomaly samples is expected to be unknown, with a large gap in anomaly regions. In order to enhance the representation ability, many methods have paid a lot of efforts in the design of network structure and loss function. Mei \textit{et al}. \cite{MSCDAE} proposed MSCDAE, which uses a Gaussian pyramid structure to obtain different receptive fields to generate more realistic background images. Yang \textit{et al}. \cite{MSFCAE} proposed a fully convolutional AE based on multi-scale feature clustering to reconstruct image backgrounds. These AE-based models train the model by minimizing the mean absolute error (L1) or squared Euclidean error (L2) between the original image and the reconstructed image, ignoring the structural information of the image, which easily leaded to blurred reconstructed images. Therefore, Bergmann \textit{et al}. \cite{AE-SSIM} proposed AE-SSIM, which applies structural similarity to an auto-encoder for image background to inspect defects. However, due to the strong generalization ability of the neural networks, some defects will also be perfectly reconstructed. And none of the above methods have the ability to deal with abnormal features properly. To solve this problem, MemAE \cite{MemAE} proposes a memory-augmented autoencoder to improve the performance of the autoencoder by obtaining reconstruction from selected memory contents records of the normal data. MemAE can repair the abnormal samples well, amplify reconstruction errors in abnormal areas, which strength the reconstruction error as the anomaly detection criterion. Recently, many improved methods based on memory mechanisms have been proposed. TrustMAE \cite{TrustMAE} presents a noise-resilient and unsupervised framework for product defect detection. Hou \textit{et al}. \cite{divide} explained image reconstruction from the perspective of feature map division and assembly, and thus proposed a block-wise memory module. However, from the experimental results, MemAE and its variants repair the abnormal foreground well but blur the normal background at the same time.

Recently, GANs \cite{GANs} have played an important role in the field of style transfer \cite{style-transfer} and image generation \cite{image=generation}, which shows that GANs have a strong generative ability. Schlegl \textit{et al}. \cite{AnoGan} firstly applied GANS in the field of anomaly detection. Since the original GANs lacks the mapping from the image domain to the latent feature domain, f-AnoGAN \cite{f-anogan}, AAE \cite{AAE}, OCGAN \cite{OCGAN}, and 
GPND \cite{GPND} are proposed. Since then, many potential anomaly detection models base on GAN have been proposed. GANomaly \cite{Ganomaly} proposes an encoder-decoder-encoder framework to minimize the distance between original and the reconstructed image in both image space and latent feature space. In order to increase the details of the reconstructed image, Skip-GANomaly \cite{skip-ganomaly} combines Skip-connection and GANomaly to reconstruct a more realistic image background. However, due to the introduction of the encoder-decoder networks structure, the above GANs-based methods are also unable to properly handle abnormal features. Therefore, Yang \textit{et al}. \cite{AFEAN} proposed AFEAN to eliminate the effect of defect reconstruction by editing defect sample features. Niu \textit{et al}. \cite{niu} proposed a memory-augmented adversarial autoencoder for defect detection, which edits the latent features through memory mechanism and ConvLSTM \cite{ConvLSTM}. In general, most of these methods are trained only on positive samples, and localize defects through residuals between original and reconstructed images. However, these methods do not achieve good performance because they exploit the pixel gap between the original image and the reconstructed image to locate defects, resulting in inaccurate defect localization and a large amount of noise.

\section{Proposed CMA-AE Methodology}
In this section, the proposed Clear Memory-Augmented Auto-Encoder (CMA-AE) is introduced in detail. Firstly, the overall network architecture is briefly introduced. Then, the main modules of CMA-AE are divided into five parts for detailed introduction, including the clear memory-augmented module (CMAM), general artificial anomaly generation algorithm (GAAGA), the multi scale feature residual (MSFR), the Encoder and the Decoder and the two-stage training strategy. Finally, the details of the reconstruction loss and the design of the loss function are discussed.
\subsection{Overall Network Architecture of CMA-AE}
The overall structure of the CMA-AE is shown in Fig. \ref{networks-fig}. The proposed CMA-AE consists of five major components: GAAGA (for generating artificial anomaly images), Encoder (for extracting latent features from images), CMAM (for inpainting anomalous foregrounds and reconstructing normal backgrounds in latent space), Decoder (for reconstructing images from latent features) and Multi Scale Feature Residual (for anomaly segmentation).

The training phase is divided into two stages. In the first stage, our training set contains only positive samples $I_p$. Firstly, we divide the images into $k^{th}$ patches (patch-size: 64$\times$64) and extract the latent features $z_k$ by Encoder. Secondly, the latent features $z_k$ are re-encoded by CMAM to get memorized latent features $\hat{\hat{{z_k}}}$. Finally, the $\hat{\hat{{z_k}}}$ is fed into Decoder to get reconstruct images ${I_p}'$. The CMAM, Encoder, and Decoder are optimized simultaneously. In the second stage, the Memory of CMAM is no longer optimized. To make the model better address abnormal features, artificial negative samples $I_{an}$ are generated by GAAGA and used as the training set. Through the forgetting and inputting mechanism of CMAM, the abnormal foreground is repaired, the normal background is reconstructed, and finally the images ${I_{an}}'$ is obtained.

During the testing phase, the negative samples $I_n$ are reconstructed by Encoder, CMAM, and Decoder to obtain the images ${I_n}'$. MSFR uses multi-scale information of input $I_n$ and output ${I_n}'$ for more accurate anomaly segmentation.


\subsection{General Artificial Anomaly Generation Algorithm}
Due to the limitations of artificial design, artificial anomaly samples usually can only simulate a few real anomalies. Inspired by Cutpaste \cite{cutpaste}, we proposed a GAAGA that cuts an image patch and pastes at a random location of an image, as shown in Fig. 3. First, based on the feature redundancy of natural images relative to industrial images, we randomly crop a 256$\times$256 patch from a natural image in the ImageNet dataset \cite{ImageNet}. Then, to more closely simulate anomalous features, we also randomly crop the patch from normal samples and resize them to 256$\times$256 before blurring. In particular, blurring allows patches to be considered degenerate anomalies, and resizing enables the networks to learn not the ability to deblur, but the ability to repair anomalies. Finally, the patches obtained in steps 1 and 2 are randomly pasted into the normal samples, as follows:
\begin{equation}
    I_{an} = paste(I_p, crop(I_n), blur(resize(crop(I_p))))
    \label{eq-gaaga}
\end{equation}
where the positive samples and natural samples are represented by $I_p$ and $I_n$. The operations of random crop, resize and blur are represented by $crop()$, $resize()$, and $blur()$. $paste(x_1, x_2, x_3)$ represents the operation of randomly pasting $x_2$, $x_3$ onto $x_1$.

\begin{figure}[!t]
    \centering
    \includegraphics[width=87mm]{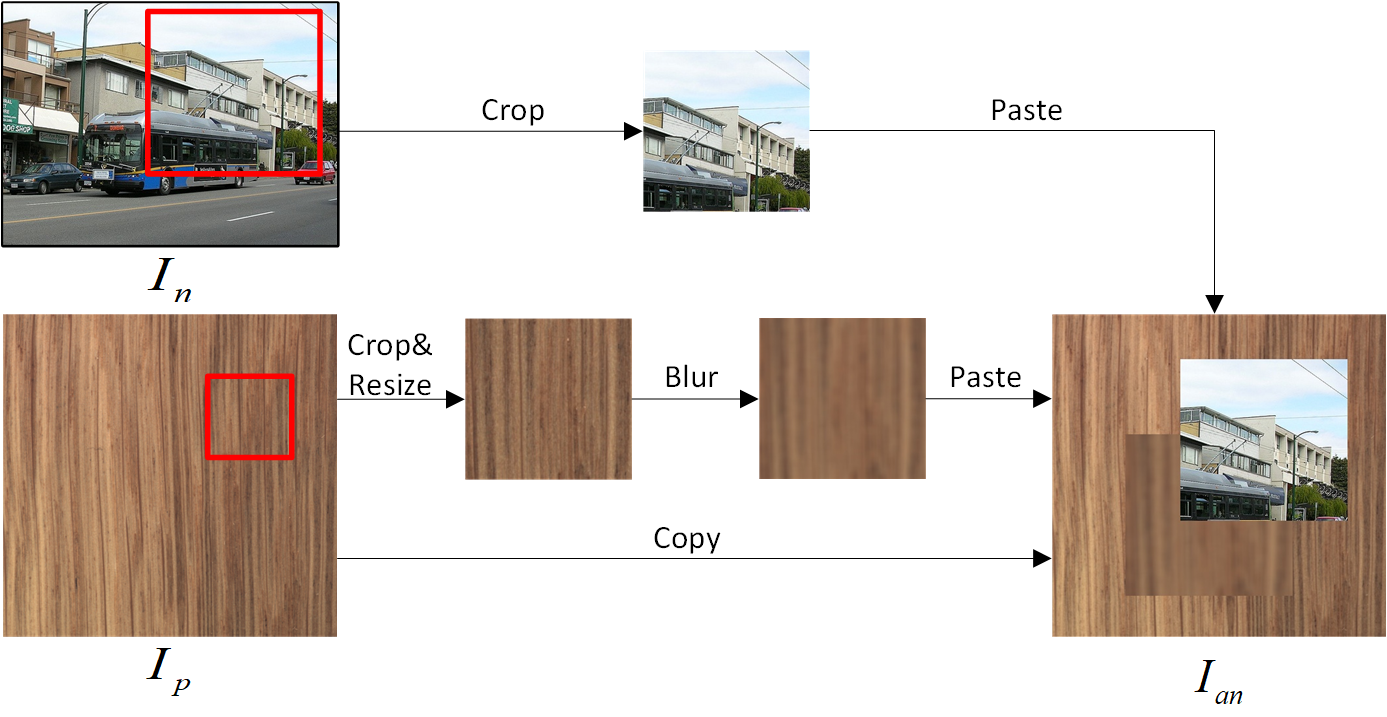}
    \caption{General artificial anomaly generation algorithm. The artificial negative samples are obtained by Crop-and-Paste of natural images and positive samples.}
    \label{fig:my_gaaga}
\end{figure}

\begin{table}
\caption{CMA-AE Architecture}
\label{table:cma-ae-architecture}
\setlength{\tabcolsep}{3pt}
\begin{threeparttable}
\begin{tabular}{p{\columnwidth}}
$\includegraphics[width=\columnwidth]{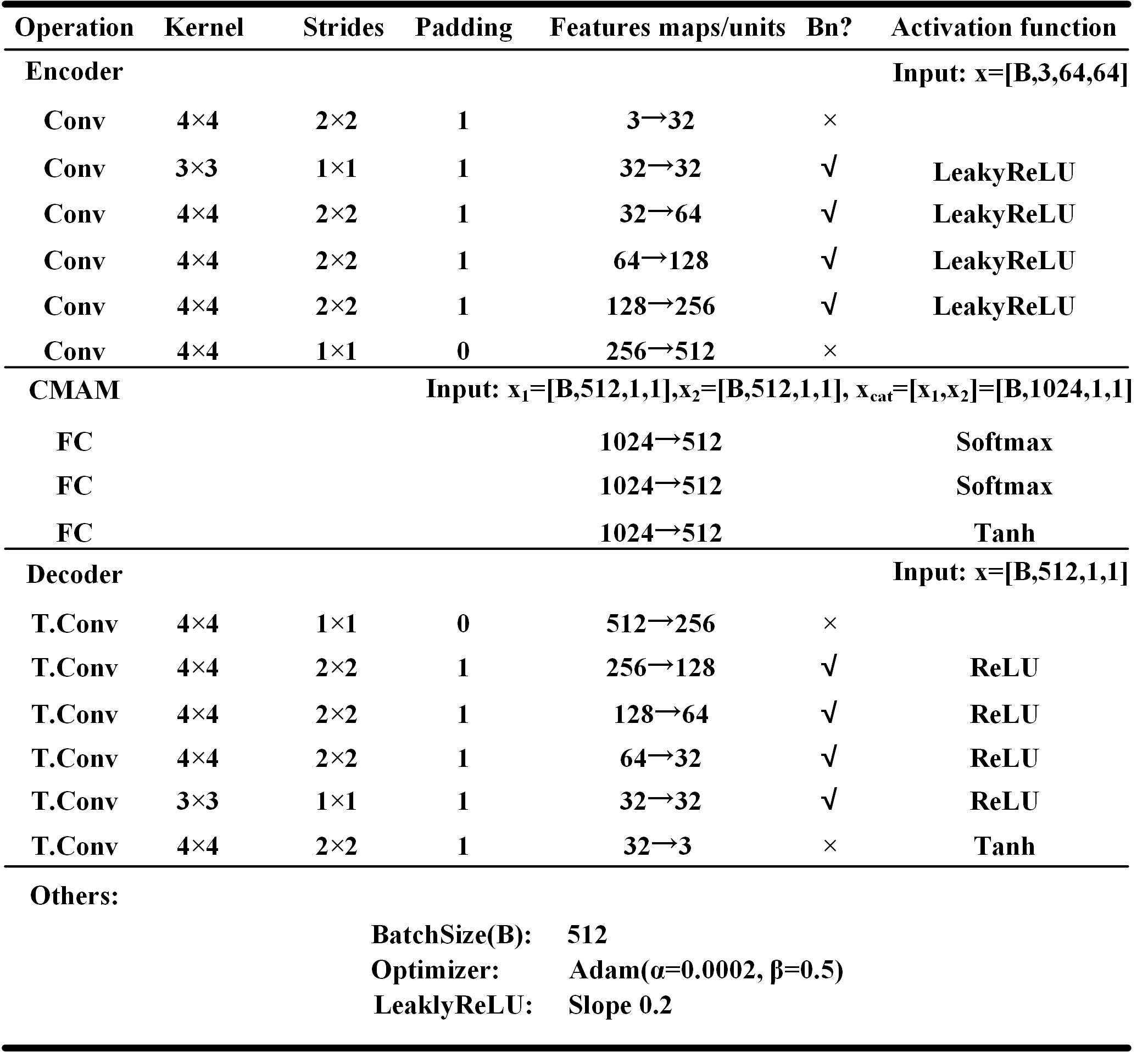}$
\end{tabular}
\begin{tablenotes}
    \footnotesize
    \item[1] Conv, FC, and T.Conv represent the convolution, full connect, and transposed convolution operations, respectively.
\end{tablenotes}
\end{threeparttable}
\end{table}

\begin{figure}
    \centering
    \includegraphics[width=87mm]{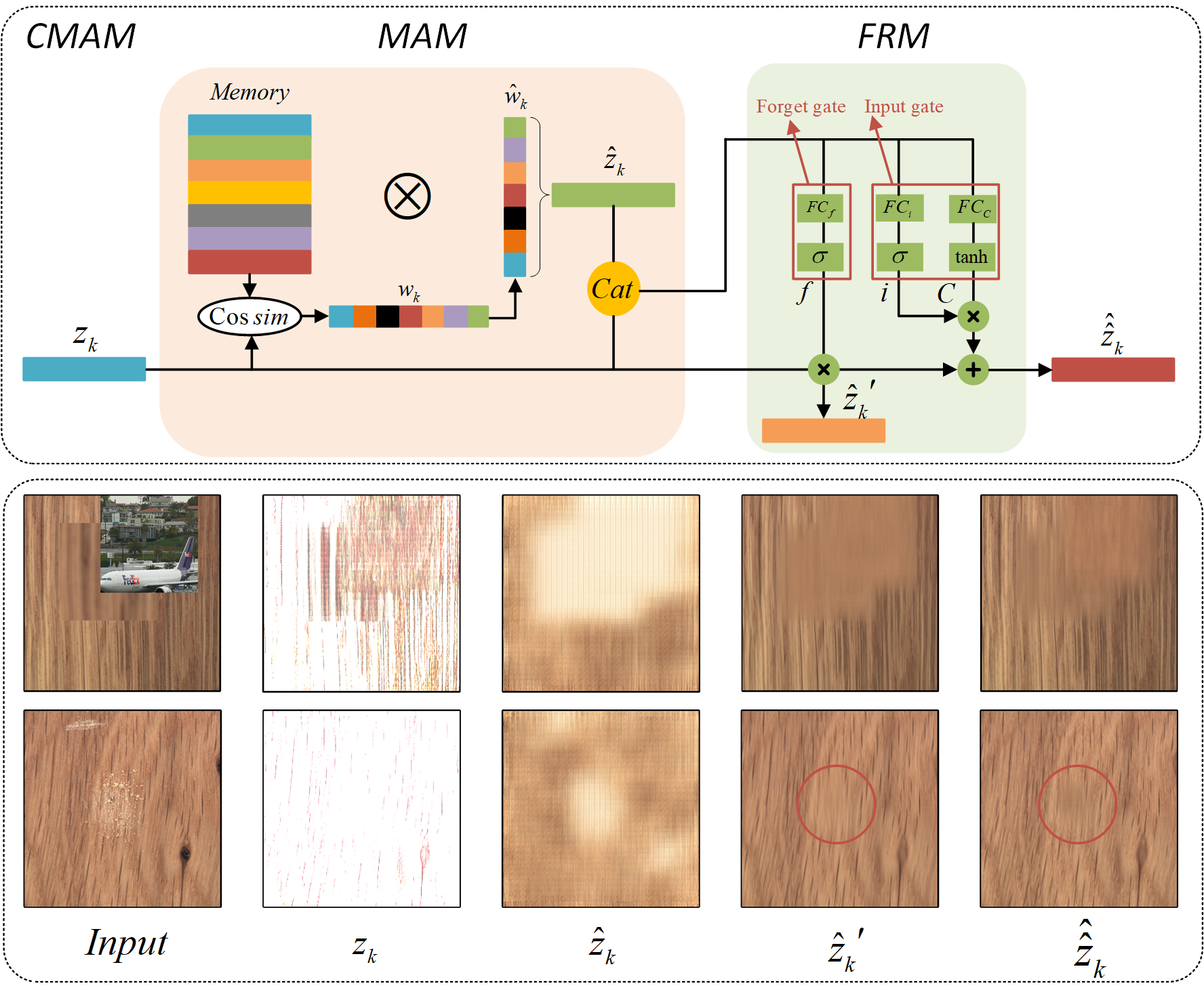}
    \caption{Diagram of the proposed clear memory-augmented module (CMAM). The CMAM includes a memory-augmented module (MAM) and a feature repair module (FRM). And we show the image obtained by the decoder at each stage of encoding.}
    \label{fig:my_cmam}
\end{figure}

\subsection{Encoder and Decoder}
In the reconstruction task, the images are mapped to the feature space by the encoder and then mapped to the image space by the decoder. By setting an information bottleneck, a dimensionality-reduced data representation is obtained in the feature space. Considering the representation ability and computational cost, we design symmetric encoders and decoders, as depicted in Table \ref{table:cma-ae-architecture}.
In the encoder, 4$\times$4 convolution kernels with a 2$\times$2 strides are used for dimensionality reduction. In order to obtain a larger receptive field, the second layer of the encoder adopts 3$\times$3 convolution kernels with strides of 1$\times$1. Symmetric, the decoder uses 4$\times$4 deconvolution kernels with a 2$\times$2 strides, and the penultimate layer uses 3$\times$3 deconvolution kernels with strides of 1$\times$1.

\subsection{Clear Memory-Augmented Module}
In order to make the network capable of repairing abnormal foreground and reconstructing normal background at the same time, based on MemAE \cite{MemAE}, we propose a clear memory-augmented module (CMAM). As shown in Fig. \ref{fig:my_cmam}, the CMAM includes a memory-augmented module (MAM) and a feature repair module (FRM). With MAM, we can get de-anomaly coding $\hat{z_k}$ but lose details. But at the same time, the coding $z_k$ obtained by encoder is rich in details. Therefore, FRM is proposed, which uses forget gates to erasure anomalous foregrounds in coding to get ${\hat{z_k}}'$. And then we use the memory-encoded information to repair the erased features through input gates to obtain $\hat{\hat{z_k}}$. Furthermore, to elucidate the effectiveness of the proposed method, we feed the encoding obtained at each step into the decoder to obtain the corresponding image.
\subsubsection{Memory-Augmented Module} \label{MAM}
In MAM, the encoding from the encoder is used as a query to retrieve the most relevant memory items from the memory bank for decoding to get de-anomaly coding, as shown in Fig. \ref{fig:my_cmam}.

The memory bank is used to record the prototypical normal patterns during training, implemented as a matrix $M\in{R^{N{\times}C}}$, where $N$ represents the number of memory items, and $C$ represents the fixed dimension. Specifically, we set $C$ equal to the dimension of the latent feature vector $z_k\in R^{1\times C}$, which denotes latent feature of the $k^{th}$ patch in the input image. And $m^i\in R^{1\times C}(i\in \{1,2,\dots,N\})$ is used to refer to the $i^th$ memory item of memory bank $M$. We define the memory bank as a content addressable memory \cite{weston2015memory,scaling-memory} with a specific addressing scheme.

The memory-coding $\hat{z_k}$ is addressed by the attention weight $w_k\in R^{N\times 1}$, and $w_k^i(i\in \{1,2,\dots,N\})$ is obtained by computing the cosine similarity between $k^{th}$ input latent feature $z_k$ and $i^{th}$ memory item $m^i$ in the memory bank, as follows:
\begin{equation}
w_k^i = \frac{{\exp (\frac{{{z_k}{{({m^i})}^T}}}{{\left\| {{z_k}} \right\|\left\| {{m^i}} \right\|}})}}{{\sum\nolimits_{j = 1}^N {\exp (\frac{{{z_k}{{({m^j})}^T}}}{{\left\| {{z_k}} \right\|\left\| {{m^j}} \right\|}})} }}
\label{eq-2}
\end{equation}
where $\|\cdot\|$ denotes the modulus length of vector. Additionally, items with attention weights less than $1/N$ are removed and re-normalized, as follows:
\begin{equation}
    \hat w_k^i = \frac{{\max (w_k^i - \frac{1}{N},0) \cdot w_k^i}}{{|w_k^i - \frac{1}{N}| + \varepsilon }}
    \label{eq-3}
\end{equation}
where $max(\cdot,0)$ is ReLU activation function, and $\varepsilon$ is a very small positive scalar. After the shrinkage operation, we re-normlizaed $\hat{w_k}$ by letting $\hat w_k^i = \frac{{\hat w_k^i}}{{\left\| {{{\hat w}_k}} \right\|}}$. Then memory-coding $\hat{z_k}$ is computed as follows:
 \begin{equation}
     {\hat z_k} = {\hat w_k}M = \sum\nolimits_{i = 1}^N {\hat w_k^i{m^i}} 
     \label{eq-4}
 \end{equation}

As suggested in \cite{MemAE}, sparse loss function is leveraged to further improve the sparsity of the attention weight $\hat{w}$:
\begin{equation}
{L_s} = \sum\nolimits_{k = 1}^K {\sum\nolimits_{i = 1}^N { - \hat w_k^i \cdot \log (\hat w_k^i)} } 
    \label{eq-5}
\end{equation}
This sparse loss function in Eq. \ref{eq-5} and the shrinkage operation in Eq. \ref{eq-3} jointly improve the sparsity of the attention weights.
\subsubsection{Feature Repair Module}
The coding $z_k$ obtained by the encoder and the de-anomaly coding $\hat{z_k}$ addressed by MAM are complementary in features, where the former is rich in texture but with defective information and the latter is de-anomaly but less rich in texture. Therefore, we concatenate $z_k$ and $\hat{z_k}$ to get $[z_k,\hat{z_k}]$. The process of combining coding $z_k$ and memory coding $\hat{z_k}$ is represented in Eqs. \ref{eq-6}-\ref{eq-10}. At first, the forget gate utilizes the information of memory coding $\hat{z_k}$ to erasure anomalous foregrounds in the coding $z_k$, as shown in Eq. \ref{eq-6}, Then, as shown in Eqs. \ref{eq-7} and \ref{eq-8}, the rich texture information is procured by input gate. Finally, the de-anomaly and richly textured features features are obtained by using Eqs. \ref{eq-9} and \ref{eq-10}.

Forget Gate:
\begin{equation}
f = \sigma ({W_f} \cdot [{z_k},{\hat z_k}] + {b_f})
\label{eq-6}
\end{equation}

Input Gate:
\begin{equation}
i = \sigma ({W_i} \cdot [{z_k},z] + {b_c})
\label{eq-7}
\end{equation}
\begin{equation}
    C = \tanh ({W_c} \cdot [{z_k},z] + {b_c})
    \label{eq-8}
\end{equation}

Forget-and-input update:
\begin{equation}
    {{\hat z_k}'} = f \cdot {\hat z_k}
    \label{eq-9}
\end{equation}
\begin{equation}
    {\hat {\hat {z_k}}} = {\hat z_k} + i \cdot C
    \label{eq-10}
\end{equation}
where $f,i,C$ are 2-D tensors, $\sigma(\cdot)$ denotes softmax function, $tanh(\cdot)$ is tanh activation function, $W_f, W_i, W_c$ are the weights of fully connection layers and $b_f, b_i, b_c$ are the biases of fully connection layers.

As shown in Fig. \ref{fig:my_cmam}, the reconstructed image by the coding ${\hat{z_k}}'$ through forget gate is less defective. Based on the forget gate, the reconstructed image by the coding $\hat{{\hat{z_k}}}$ through the input gate is richer in texture.

\begin{figure}[!t]
    \centering
    \includegraphics[width=87mm]{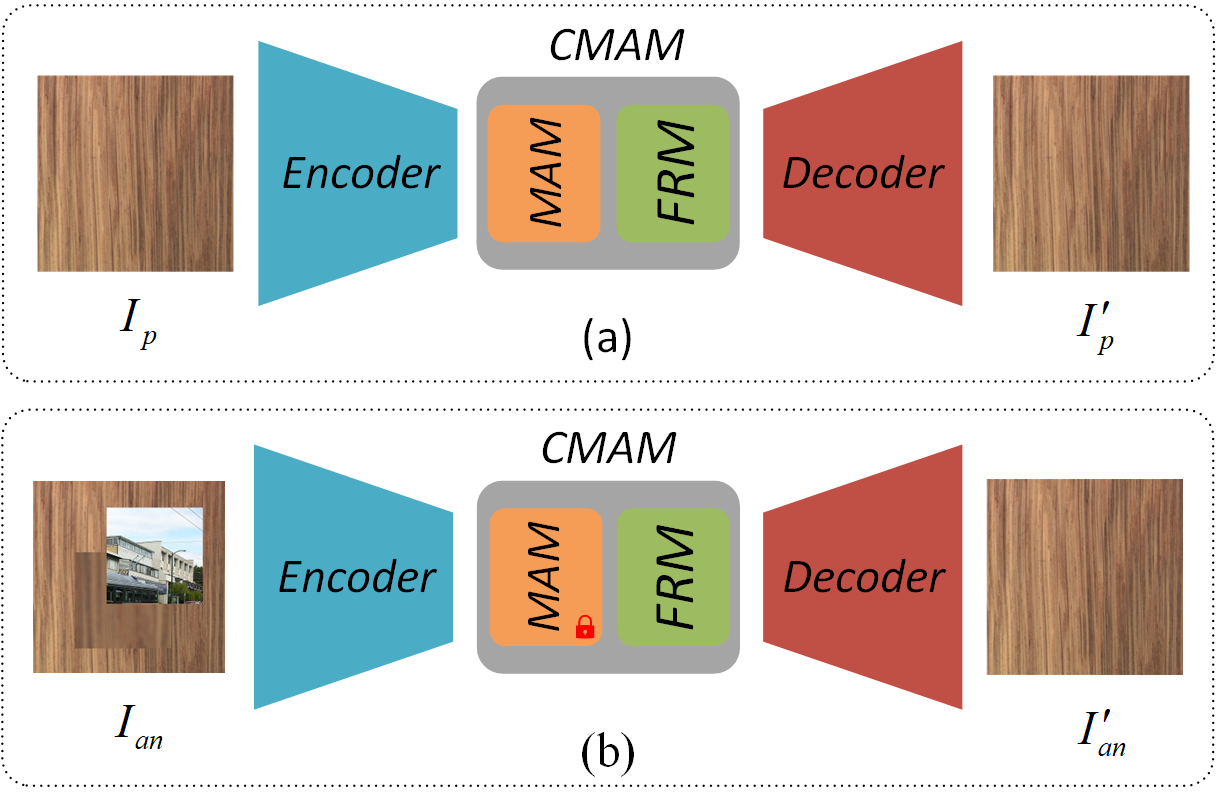}
    \caption{Illustration of the two-stage training strategy of CMA-AE. (a) Training stage 1, trained with positive samples, the Encoder, Decoder, MAM, and FRM are updated together. (b) Training stage 2, trained with artificial negative samples, MAM is no longer updated.}
    \label{fig:two-stage}
\end{figure}

\subsection{Two-stage Training Strategy}
As discussed above, it is difficult to make the model have the ability to reconstruct normal background and repair abnormal foreground at the same time in one stage of training. Therefore, we propose a two-stage training strategy, as shown in Fig. \ref{fig:two-stage}. In the first procedure, the model is optimized by training positive samples so that the MAM record prototypical patterns of normal features. In the second procedure, the memory bank in MAM is fixed, and the Encoder, Decoder, and FRM are optimized by training artificial negative samples so that the FRM learns how to forget defective information and input normal textured information. As  shown in Fig. \ref{fig:two-stage} (a), in training stage 1, considering that the model optimized with L2-norm will produce blurry reconstructed images, L1-norm is used as the criterion for measuring distance:
\begin{equation}
     \centering
      L_{rec1} = \underset{I_p\sim P_{I_p}}{\mathbb{E} }\left [{||I_p-{I_p}'||}_1\right ]  
      \label{eq:L-rec1}
\end{equation}
where $||\cdot||_1$ denotes the L1-norm. To facilitate the sparsity of the attention weights, as described in Section \ref{MAM}, CMA-AE is trained with the sparsity loss $L_s$. Therefore, the overall model is optimized under the joint loss function:
\begin{equation}
    \centering
    L_1 = w_1L_{rec1}+w_2L_s
\end{equation}
where $w_1, w_2$ are the weights that control the relative importance of two terms. In this paper, we recommend setting $w_1$=50 and $w_2$=0.01.

As shown in Fig. \ref{fig:two-stage} (b), in training stage 2, same as training stage 1, L1-norm is used as the criterion for measuring distance:
\begin{equation}
     \centering
      L_{rec2} = \underset{I_{an}\sim P_{I_{an}}}{\mathbb{E} }\left [{||I_{an}-{I_{an}}'||}_1\right ]  
      \label{eq:L-rec2}
\end{equation}
MAM is fixed, Encoder, FRM and Decoder are optimized under the following joint loss function:
\begin{equation}
    \centering
    L_2 = w_1L_{rec2}+w_2L_s
\end{equation}
The values of $w_1, w_2$ are same as in the training stage 1. Finally, we summarize the two-stage training strategy in Algorithm \ref{alg:two-stage}.
\begin{algorithm}[!t]
    \caption{Two-stage training strategy}
    \label{alg:two-stage}
    \SetKwInOut{KwOut}{Parameters}
     \KwData{A generated dataset composed of positive images $I_p$ and artificial negative images $I_{an}$}
     \KwIn{Encoder \textit{\textbf{E}}, Decoder \textit{\textbf{D}}, \textit{\textbf{CMAM}} (\textit{\textbf{MAM}}, \textit{\textbf{FRM}})}
    \KwOut{Iterations $T_1$ and $T_2$, learning rate $\alpha$}
    Training stage 1:\\
    \For{$i=1, 2, ..., T_1$}
    {
    1. Extract patches from $I_p$ \;
    2. Calculate partial derivatives: \\
       \quad $\nabla_\textit{\textbf{E}}L_1$, $\nabla_\textit{\textbf{D}}L_1$, $\nabla_\textit{\textbf{CMAM}}L_1$ \;
    3. Superpose partial derivatives and renew models \\
      \quad $\textit{\textbf{E}} \leftarrow \textit{\textbf{E}}+\alpha\times\nabla_\textit{\textbf{E}}L_1$\\
       \quad $\textit{\textbf{CMAM}} \leftarrow \textit{\textbf{CMAM}}+\alpha\times\nabla_\textit{\textbf{CMAM}}L_1$\\
       \quad $\textit{\textbf{D}} \leftarrow \textit{\textbf{D}}+\alpha\times\nabla_\textit{\textbf{D}}L_1$\;
       }
    Training stage 2: \\
      \For{$i=1, 2, ..., T_2$}
    {
    1. Fixed the weights of \textit{\textbf{MAM}} \;
    2. Extract patches from $I_{an}$ \;
    3. Calculate partial derivatives: \\
       \quad $\nabla_\textit{\textbf{E}}L_1$, $\nabla_\textit{\textbf{D}}L_1$, $\nabla_\textit{\textbf{FRM}}L_1$ \;
    4. Superpose partial derivatives and renew models \\
      \quad $\textit{\textbf{E}} \leftarrow \textit{\textbf{E}}+\alpha\times\nabla_\textit{\textbf{E}}L_1$\\
       \quad $\textit{\textbf{FRM}} \leftarrow \textit{\textbf{FRM}}+\alpha\times\nabla_\textit{\textbf{FRM}}L_1$\\
       \quad $\textit{\textbf{D}} \leftarrow \textit{\textbf{D}}+\alpha\times\nabla_\textit{\textbf{D}}L_1$\;
       }
\end{algorithm}

\subsection{Multi Scale Feature Residual}
\begin{figure}[!t]
    \centering
    \includegraphics[width=87mm]{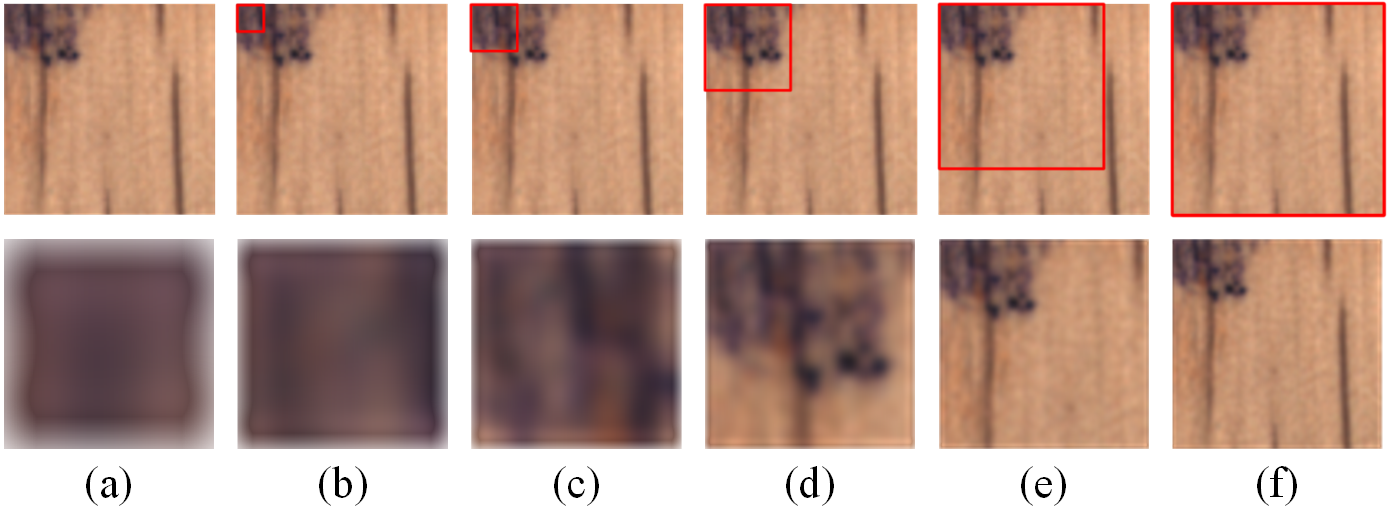}
    \caption{Different receptive fields of different scale. The red boxes indicate the region of receptive field.}
    \label{fig:receptive-field}
\end{figure}

\begin{figure}[!t]
    \centering
    \includegraphics[width=87mm]{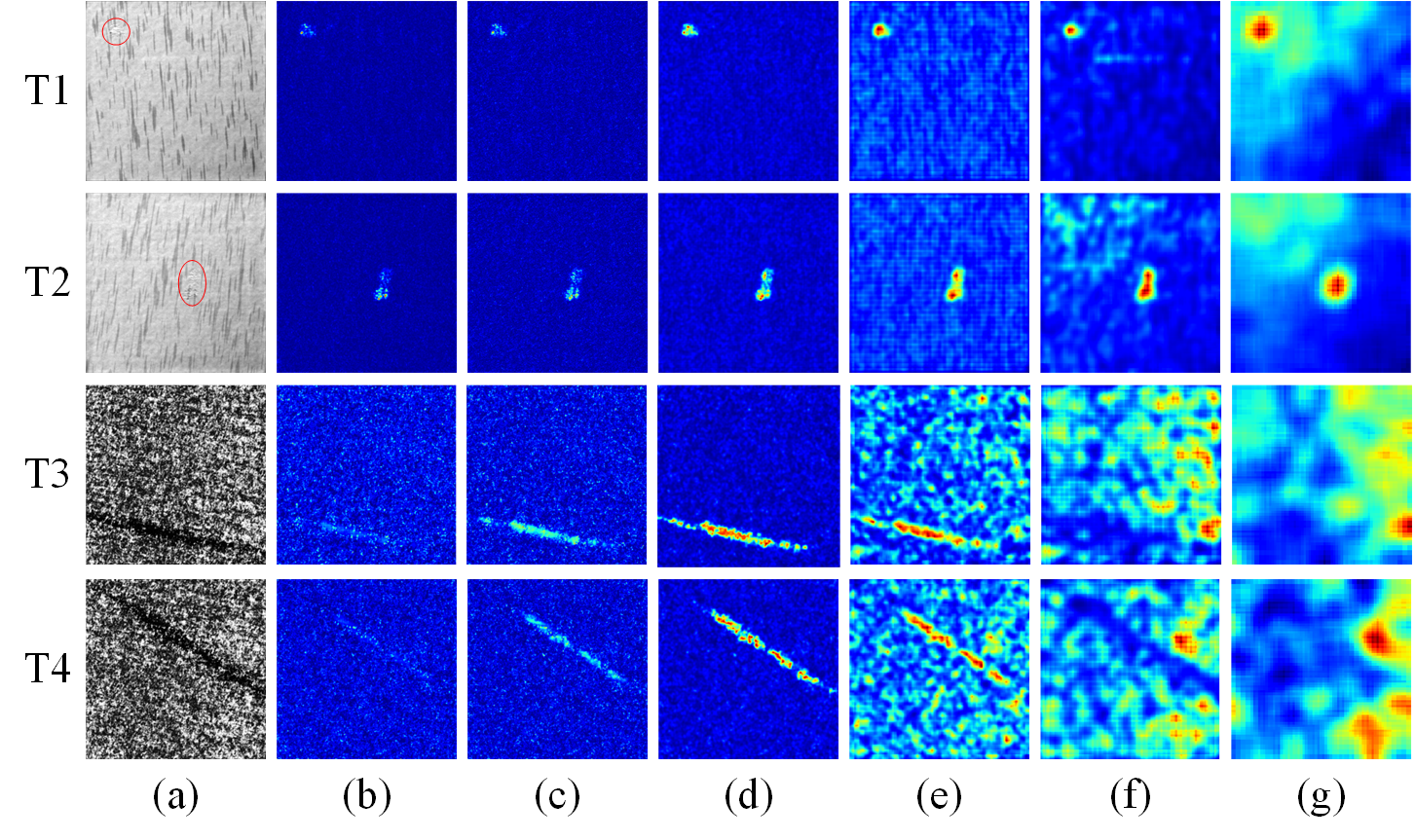}
    \caption{ Feature residual by the proposed MSFR. (a) are defective images, (b), (c), (d), (e), (f), and (g) are the error maps obtained by scale 1, scale 2, scale 3, scale 4, scale 5 and scale 6.}
    \label{fig:select_layer}
\end{figure}

In fact, anomalies are composite representations of multiple pixels with contextual relationships and have significant multi-scale properties. Therefore, to obtain accurate and noise-free anomaly segmentation maps, we propose multi scale feature residual (MSFR). At first, we utilize the residual of feature maps instead of residual of images, which guarantees the correlation between pixels. An element in the feature map corresponds to a region in the original image, which can be thought of as an anomaly score for that region, as shown in Fig. \ref{fig:receptive-field}. However, as the receptive field expands, an element in the feature map corresponds to a larger area of the image. When there are abnormal areas and normal areas exist at the same time, there will be errors whether the element is represented as normal or abnormal. To address this trade-off, we utilize multiple feature residuals to obtain multi scale information.
In many existing methods \cite{PGANet,TrustMAE}, VGG is trained as a feature extraction network on large-scale datasets, such as ImageNet \cite{ImageNet}. These networks generalize well, but do not work for specific data. In this paper, the Encoder of CMA-AE can be regarded as a feature extractor learned from self-supervised learning on a specific dataset. As shown in Fig. \ref{networks-fig}, in MSFR, when testing, the images and the corresponding reconstructions are fed to the Encoder, so we can get the multi-scale feature maps. However, not all feature maps can clearly express abnormal regions. We perform some experiments and analysis on the selection of feature maps. As illustrated in table \ref{table:cma-ae-architecture}, the Encoder has a total of 6 convolutional layers. As depicted in Fig. \ref{fig:select_layer}, the error maps of scale 4, 5, 6 have a lot of noise. In this paper, we select the feature maps of scale 1, 2, 3 to obtain abnormal segmentation results. 

Then, the feature maps subtracted and squared, upsampled to the size of the original image, and averaged over the channel dimension. Finally, a weighted average of feature residuals at multiple scales is performed. The images is represented by $I_n,{I_n}'\in R^{C\times H\times W}$, the feature maps of scale 1, 2, 3 is represented by $F_1,{F_1}'\in R^{C_1\times H_1\times W_1}$, $F_2,{F_2}'\in R^{C_2\times H_2\times W_2}$, $F_3,{F_3}'\in R^{C_3\times H_3\times W_3}$, and the corresponding receptive field is represented by ${RF}_1\in R^{{C_1}'\times {H_1}' \times {W_1}'}$, ${RF}_2\in R^{{C_2}'\times {H_2}' \times {W_2}'}$, ${RF}_3\in R^{{C_3}'\times {H_3}' \times {W_3}'}$. Therefore, $MSFR(I_n, {I_n}')\in R^{H\times W}$ are defined as follows:
\begin{equation}
\begin{split}
        MSFR({I_n},{I_n}') = (\sum\nolimits_{i = 1}^3 {({H_i}' \times {W_i}') \times \delta (\tau ({{({F_i} - {F_i}')}^2})))/} \\ (\sum\nolimits_{i = 1}^3 {({H_i}' \times {W_i}')})
\end{split}
\end{equation}
where $\delta(\cdot)$ is the operation that takes the mean in the channel dimension, and $\tau(\cdot)$ is a bilinear up-sampling function that resizes the feature tensors to $H\times W$.

\begin{table*}[!t]
\centering
\caption{The AuROC results of different methods on five types of textured surfaces in MV-TAD dataset}
\label{table:mvtad}
\begin{threeparttable}
\begin{tabular}{cccccccccccc}
\Xhline{1.5pt}
Category & AE-SSIM & AnoGAN & f-AnoGAN & {\color[HTML]{333333} MS-FCAE} & MemAE & {\color[HTML]{333333} TrustMAE}       & {\color[HTML]{333333} RIAD}           & VAE   & {\color[HTML]{333333} ACDN}        & AFEAN & \textbf{CMA-AE} \\ \hline
Carpet   & 87.00   & 54.00  & 66.00    & 78.20                          & 81.16 & {\color[HTML]{333333} \textbf{98.53}} & {\color[HTML]{333333} {\ul 96.30}}    & 73.50 & {\color[HTML]{333333} 91.10}       & 90.30 & 91.25           \\
Grid     & 94.00   & 58.00  & 85.00    & 88.10                          & 95.56 & {\color[HTML]{333333} 97.45}          & {\color[HTML]{333333} {\ul 98.80}}    & 96.10 & {\color[HTML]{333333} 94.10}       & 92.60 & \textbf{99.31}  \\
Leather  & 78.00   & 64.00  & 83.00    & 91.70                          & 92.91 & {\color[HTML]{333333} 98.05}          & {\color[HTML]{333333} \textbf{99.40}} & 92.50 & {\color[HTML]{333333} 98.40}       & 96.10 & {\ul 99.13}     \\
Tile     & 59.00   & 50.00  & 72.00    & 53.20                          & 70.76 & {\color[HTML]{333333} 82.48}          & {\color[HTML]{333333} 89.10}          & 65.40 & {\color[HTML]{333333} {\ul 93.60}} & 85.70 & \textbf{98.82}  \\
Wood     & 73.00   & 62.00  & 74.00    & 81.20                          & 85.44 & {\color[HTML]{333333} 92.62}          & {\color[HTML]{333333} 85.80}          & 83.80 & {\color[HTML]{333333} {\ul 92.90}} & 92.20 & \textbf{96.96}  \\ \hline
Ave.     & 78.00   & 58.00  & 76.00    & 78.50                          & 85.17 & 93.83                                 & 93.70                                 & 82.26 & {\ul 94.10}                        & 91.40 & \textbf{97.09}  \\ \Xhline{1.5pt}
\end{tabular}
\begin{tablenotes}
\footnotesize  
\item[1] The best AuROC result is in bold, and the second best is underlined.
\end{tablenotes}
\end{threeparttable}
\end{table*}

\begin{figure*}[!t]
    \centering
    \includegraphics[width=180mm]{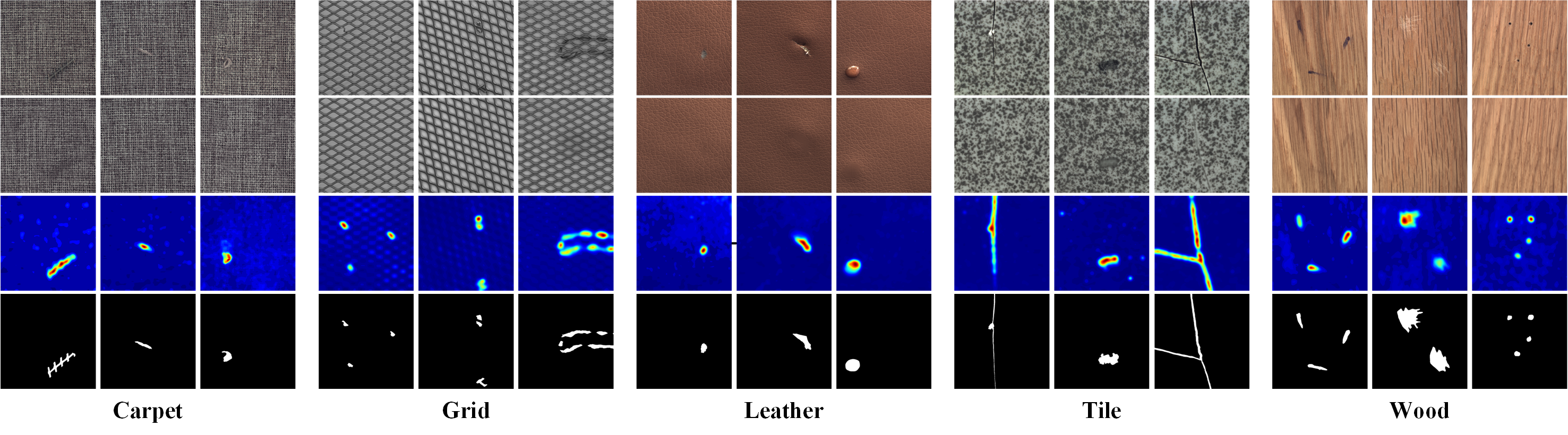}
    \caption{Inspection results for five types of textures in MV-TAD \cite{MVTEC}. From top to bottom are the input defective images, the reconstruction images, the error maps and the ground truth, respectively.}
    \label{fig:mvtad-examples}
\end{figure*}

\section{Experimentation}
\subsection{Set Up}
In this section, we analyze the performance of the proposed CMA-AE method through several sets of experiments:
\begin{itemize}
    \item The overall inspection performance of CMA-AE is compared with nine state-of-the-art methods on five benchmark textured surfaces in the MV-TAD dataset \cite{MVTEC}.
    \item To further verify the generalizability of the model, a comparative experiment on four challenging textured surfaces in the DAGM dataset \cite{DAGM} is conducted.
    \item The influences of each component in the CMA-AE are explored in ablation experiments.
    \item The inference speed of CMA-AE is compared with other outstanding methods.
    \item The CMA-AE is evaluated on an industrial dataset to validate its industrial potential.
\end{itemize}

In these experiments, a variety of anomaly detection samples are used, including carpet, grid, leather, tile, wood, wallpaper, Bcement, MAGtile, and WHcement.
The carpet, grid, leather, tile, and wood textured surfaces are sourced from MV-TAD \cite{MVTEC}, and the wallpaper, Bcement, MAGtile, and WHcement textured surfaces are sourced from DAGM \cite{DAGM}. All images are resized to 512x512 pixels.

To quantitatively analyze the performance of various methods, we adopt the area under the receiver operating characteristic curve (AuROC) as evaluation criterion, which is insensitive to thresholds and can better evaluate the inspection performance of models.

All the experiments are implemented using Python 3.8.0 and Pytorch 1.9.1 on a computer with an NVIDIA Tesla A100 GPU, which is equipped with 40 Intel(R) Xeon(R) CPU E5-2640 v4 at 2.40GHz and 40GB memory.

\subsection{Overall Performance Comparison on MV-TAD dataset}
To verify the overall performance of the proposed CMA-AE method, the inspection performance of CMA-AE is compared with a variety of outstanding anomaly detection methods, including AE\_SSIM \cite{AE-SSIM}, AnoGAN \cite{AnoGan}, f-AnoGAN \cite{f-anogan}, MS-FCAE \cite{MSFCAE}, MemAE \cite{MemAE}, RIAD \cite{RIAD}, TrustMAE \cite{TrustMAE}, VAE \cite{VAE}, ACDN \cite{ACDN} and AFEAN \cite{AFEAN}.
\begin{figure}[!t]
    \centering
    \includegraphics[width=75mm]{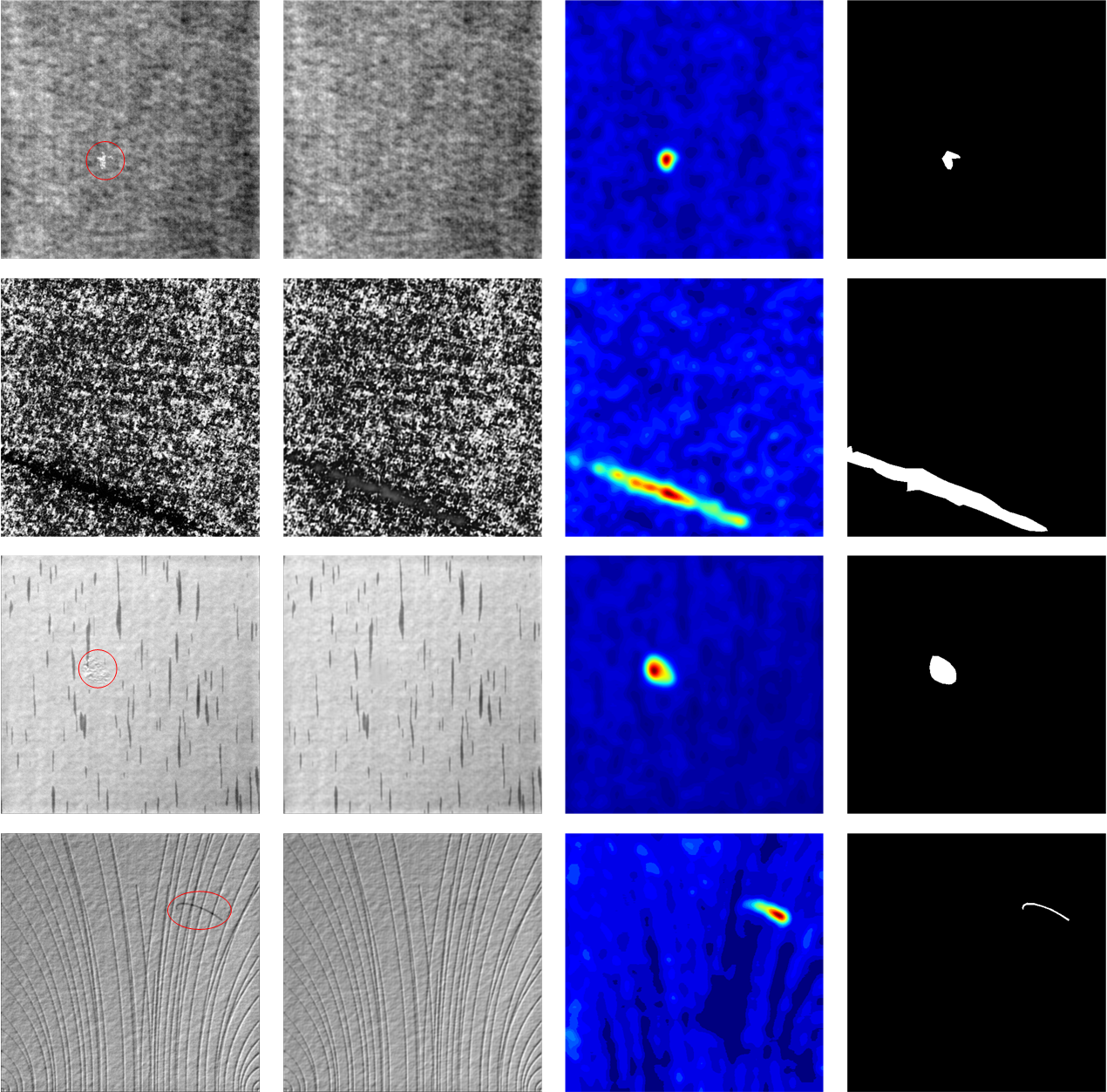}
    \caption{Inspection results for four types of textures in DAGM \cite{DAGM}. From left to right are the input defective images, the reconstruction images, the error maps and the ground truth, respectively.}
    \label{fig:dagm-examples}
\end{figure}
The experimental results of quantitative analysis are presented in Table \ref{table:mvtad}. CMA-AE achieves a better average result compared to other outstanding methods. Especially for grid, tile, and wood, compared to the second-best result, CMA-AE improves AuROC metric by margins of 0.51\%, 5.22\%, and 4.06\%, respectively. CMA-AE performs well on five different samples, which shows that our method can maintain performance on a variety of samples.

Some inspection results of CMA-AE on five different samples are shown in Fig. \ref{fig:mvtad-examples}. The CMA-AE leverages GAAGA and CMAM to enable the model can simultaneously repair anomalous foregrounds and reconstruct normal backgrounds, thus obtaining clear reconstructed images. Instead of using pixel difference between input and reconstructed images, CMA-AE adopts multi scale feature residual to obtain more accurate defect segmentation maps.

\begin{table}[!t]
\centering
\caption{The AuROC results of different methods on four types of textured surfaces in DAGM dataset}
\label{table:dagm}
\begin{threeparttable}
\begin{tabular}{ccccc|c}
\Xhline{1.5pt}
{\color[HTML]{333333} Category}        & {\color[HTML]{333333} MAGtile}        & {\color[HTML]{333333} Bcement}        & {\color[HTML]{333333} wallpaper}      & {\color[HTML]{333333} WHcement}       & {\color[HTML]{333333} Ave.}           \\ \hline
{\color[HTML]{333333} AE-SSIM}        & {\color[HTML]{333333} 74.30}          & {\color[HTML]{333333} 79.40}          & {\color[HTML]{333333} 95.60}          & {\color[HTML]{333333} 81.70}          & {\color[HTML]{333333} 82.75}          \\ \hline
{\color[HTML]{333333} CNN\_Dict}       & {\color[HTML]{333333} 69.20}          & {\color[HTML]{333333} 75.80}          & {\color[HTML]{333333} 66.50}          & {\color[HTML]{333333} 67.90}          & {\color[HTML]{333333} 69.85}          \\ \hline
{\color[HTML]{333333} AnoGAN}          & {\color[HTML]{333333} 78.10}          & {\color[HTML]{333333} 45.50}          & {\color[HTML]{333333} 72.20}          & {\color[HTML]{333333} 72.70}          & {\color[HTML]{333333} 67.13}          \\ \hline
{\color[HTML]{333333} OCGAN}           & {\color[HTML]{333333} 89.10}          & {\color[HTML]{333333} 74.50}          & {\color[HTML]{333333} 97.10}          & {\color[HTML]{333333} 95.60}          & {\color[HTML]{333333} 89.08}          \\ \hline
{\color[HTML]{333333} MS-FCAE}         & {\color[HTML]{333333} 95.90}          & {\color[HTML]{333333} 58.80}          & {\color[HTML]{333333} 90.50}          & {\color[HTML]{333333} \textbf{96.40}} & {\color[HTML]{333333} 85.40}          \\ \hline
{\color[HTML]{333333} AFEAN}           & {\color[HTML]{333333} \ul{97.40}}          & {\color[HTML]{333333} \textbf{94.80}}          & {\color[HTML]{333333} {\ul 98.30}}    & {\color[HTML]{333333} {96.10}}    & {\color[HTML]{333333} {\ul 96.65}}    \\ \hline
{\color[HTML]{333333} \textbf{CMA-AE}} & {\color[HTML]{333333} \textbf{99.97}} & {\color[HTML]{333333} \ul{90.45}} & {\color[HTML]{333333} \textbf{99.90}} & {\color[HTML]{333333} \ul{96.35}}          & {\color[HTML]{333333} \textbf{96.67}} \\ \Xhline{1.5pt}
\end{tabular}
\begin{tablenotes}
    \footnotesize
    \item[1] The best AuROC result is in bold, and the second best is underlined.
\end{tablenotes}
\end{threeparttable}
\end{table}

\subsection{Inspection Generalizability Experiment on DAGM dataset}
To further verify the generalizability of CMA-AE, the detection performance of CMA-AE is compared with various excellent methods, including AE-SSIM \cite{AE-SSIM}, CNN\_Dict \cite{CNN_Dict}, AnoGAN \cite{AnoGan}, OCGAN \cite{OCGAN}, MS-FCAE \cite{MSFCAE} and AFEAN \cite{AFEAN}.

The quantitative experimental results are shown in Table \ref{table:dagm}. CMA-AE outperforms other outstanding methods in terms of AuROC metric, which reveals that the proposed method can maintain excellent performance on different datasets, with good generalization.

Fig. \ref{fig:dagm-examples} shows some defect inspection examples of CMA-AE. The CMA-AE can inspect and locate defective regions accurately on four types of textured surfaces.
\subsection{Inference Time Comparison}
A good balance between inference speed and inspection performance is the essential point in practical industrial defect detection. To demonstrate that CMA-AE method can achieve the balance, the inference time of CMA-AE is compared with that of other outstanding methods, comprising AE-SSIM \cite{AE-SSIM}, f-AnoGAN \cite{f-anogan}, MemAE \cite{MemAE}, TrustMAE \cite{TrustMAE}, and RIAD \cite{RIAD}. The inference time is evaluated with 512$\times$512 pixels resolution.

The quantitative experimental results regarding inference time are shown in Table \ref{table:inference time}. The inference time of CMA-AE is 41.673ms, ranked behind AE-SSIM, f-AnoGAN, MemAE, and TrustMAE. However, the inspection performances of these methods are inferior to that of CMA-AE method. RIAD method 
leverages masked images at different scales to inspect defects, leading to slow inference speed, which limits its practical industrial applications. Accordingly, from the perspective of the balance between inference speed and inspection performance, the CMA-AE method is in line with industrial demands.

\begin{table}[!t]
\centering
\caption{Average Inference Time}
\setlength{\tabcolsep}{3pt}
\begin{tabular}{p{\columnwidth}}
\centering
$\includegraphics{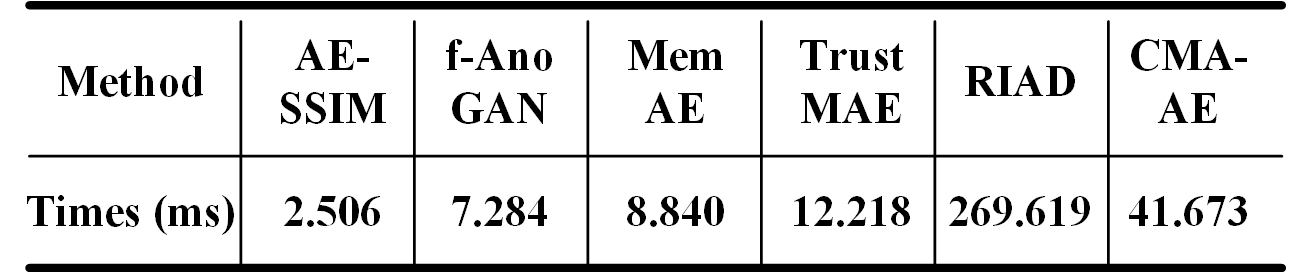}$
\end{tabular}
\label{table:inference time}
\end{table}
\begin{table}[!t]
\centering
\caption{Ablation Analysis for the CMA-AE on the Leather Dataset}
\setlength{\tabcolsep}{3pt}
\begin{tabular}{p{\columnwidth}}
\centering
$\includegraphics{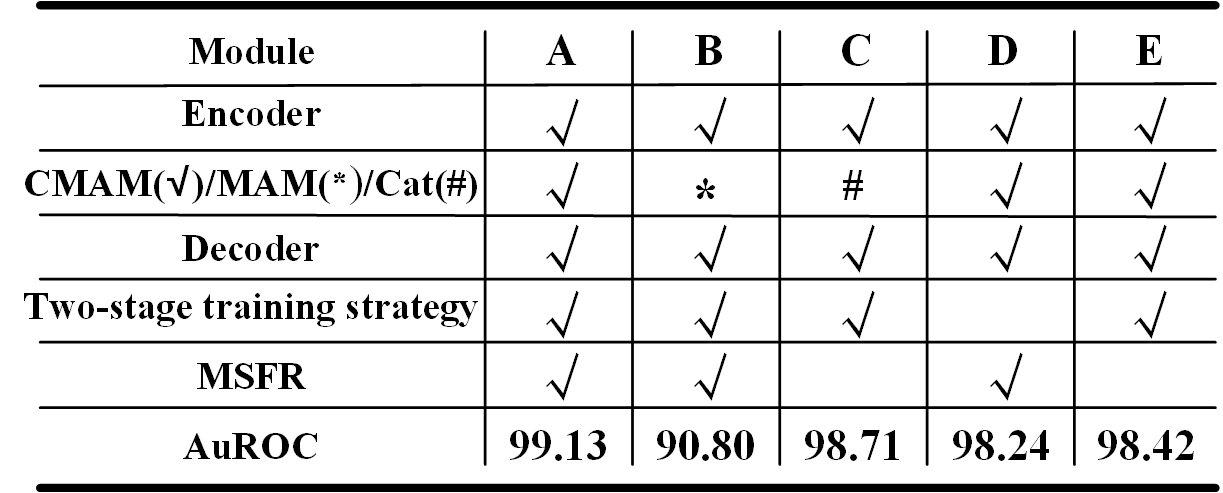}$
\end{tabular}
\label{table:ablation_analysis}
\end{table}
\begin{figure}[!t]
    \centering
    \includegraphics{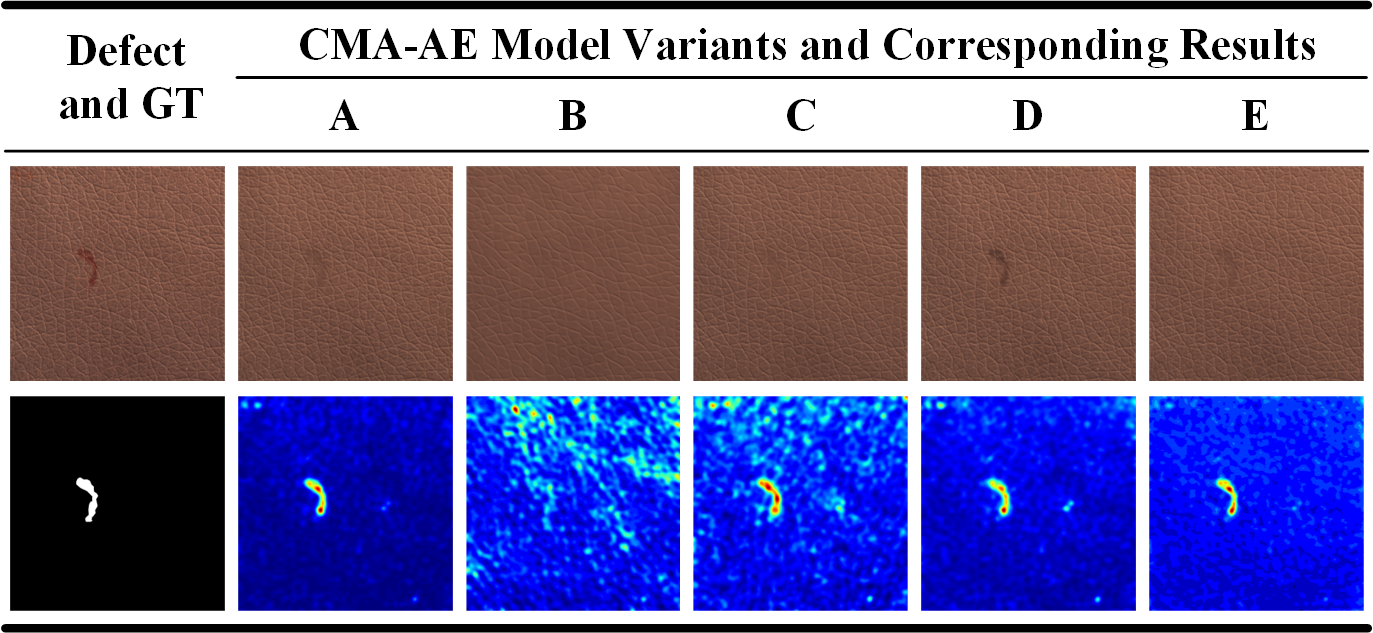}
    \caption{Examples of images from tests in the ablation study. The first row shows the reconstruct images, and the second row shows the error maps. Each column corresponds to the model variant in Table \ref{table:ablation_analysis}.}
    \label{fig:ablation_analysis}
\end{figure}
\begin{figure}[!t]
    \centering
    \includegraphics[width=75mm]{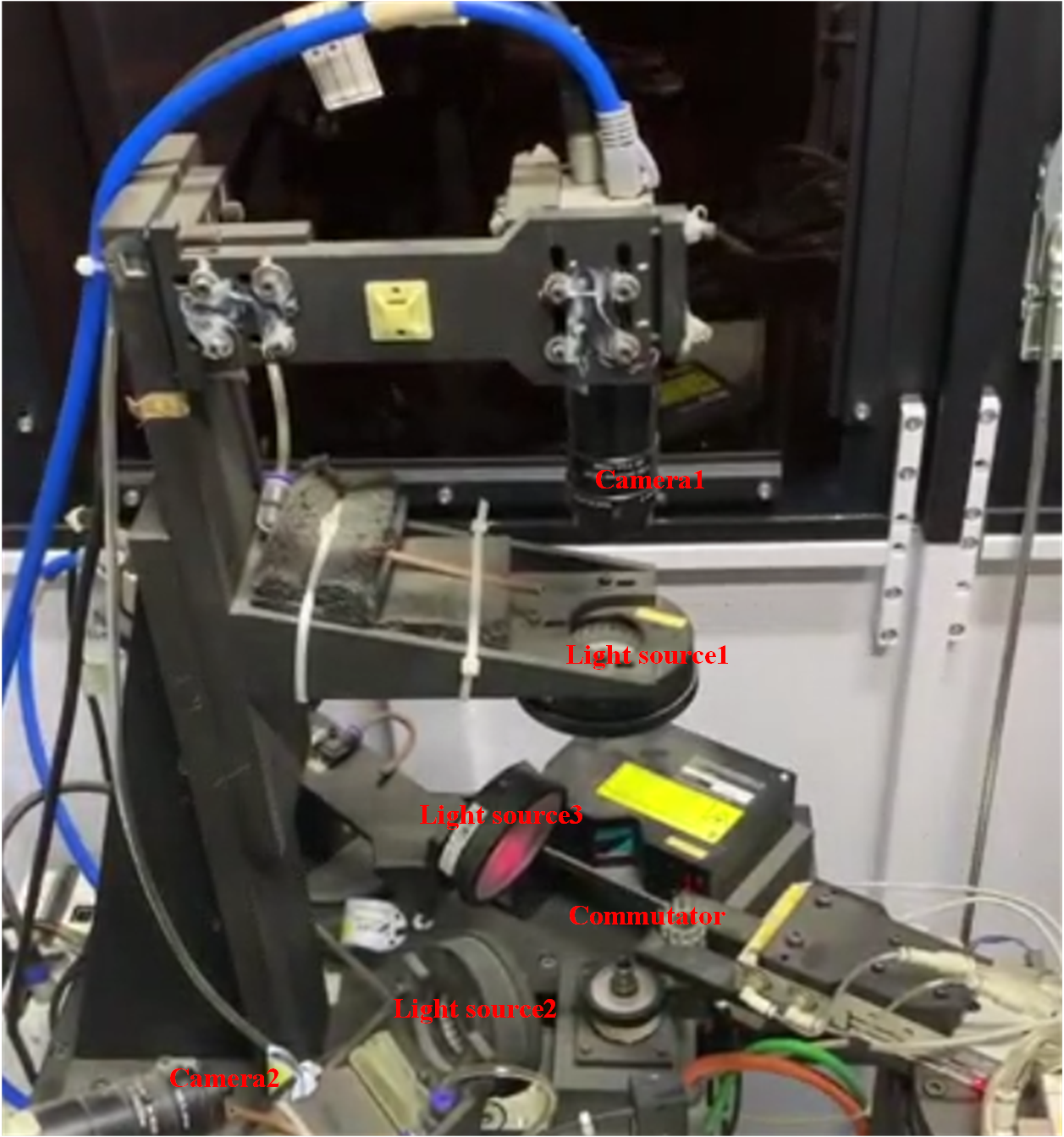}
    \caption{Automated optical inspection equipment for commutator surface defect detection.}
    \label{fig:industrial-equipment}
\end{figure}
\subsection{Ablation Analysis}
A set of ablation experiments are performed to confirm the influence of each module in CMA-AE. For a fair comparison, all evaluated variants of CMA-AE use the same parameter setting. The experimental results are shown in Table \ref{table:ablation_analysis} and Fig. \ref{fig:ablation_analysis}

\subsubsection{Influence of CMAM}
CMAM is proposed to solve the problem that MemAE \cite{MemAE} enhances the repair ability of abnormal foregrounds but weakens the ability to reconstruct normal backgrounds at the same time. To verify whether our improvement is effective, the impact of replacing CMAM with MAM will be analyzed in detail.

As shown in columns A and B of Fig. \ref{fig:ablation_analysis}, after replacing CMAM with MAM, we find that its background is not reconstructed well, resulting in a lot of noise during defect segmentation. Table \ref{table:ablation_analysis} presents the quantitative experimental results. CMA-AE (column A) improves the AuROC by a margin of 8.33\% compared to the model (column B) that replaced CMAM with MAM.

To verify that CMAM improves MAM not only due to the increase of parameters, we design a set of comparative experiments between CMA-AE and Cat (replacing forget and input gates with concatenate). As shown in columns A and C of Fig. \ref{fig:ablation_analysis}, we find that the error map of CMAM is clearer than that of Cat. Table \ref{table:ablation_analysis} shows the quantitative results. CMA-AE (column A) improves the AuROC by 0.42\% compared to the Cat (column C).

This experiment confirms that our CMAM outperforms MAM. CMAM can repair abnormal foreground through the way of forgetting and inputting and can reconstruct normal background well, helping us achieve more accurate anomaly detection.

\subsubsection{Influence of Two-stage training strategy}
The two-stage training strategy aims to let CMAM learn how to forget and input, giving the model the ability to distinguish abnormal foreground from normal background. To confirm its effectiveness, we removed the two-stage training strategy from CMA-AE.

Columns A and D of Fig. \ref{fig:ablation_analysis} show an example of the influence of 
two-stage training strategy. Without a two-stage training strategy, CMAM cannot learn how to forget and input, resulting in goon reconstruction of defects in the testing phase. The quantitative experimental results are illustrated in Table \ref{table:ablation_analysis}, compared with model without the two-stage training strategy (column D), CMA-AE (column A) improves the AuROC by a margin of 0.89\%.

The experiment confirms that our two-stage training strategy is effective. The two-stage training strategy can make the model have the ability to distinguish abnormal foregrounds from normal backgrounds, which helps the model to better repair abnormal foregrounds.

\begin{figure}[!t]
    \centering
    \includegraphics[width=75mm]{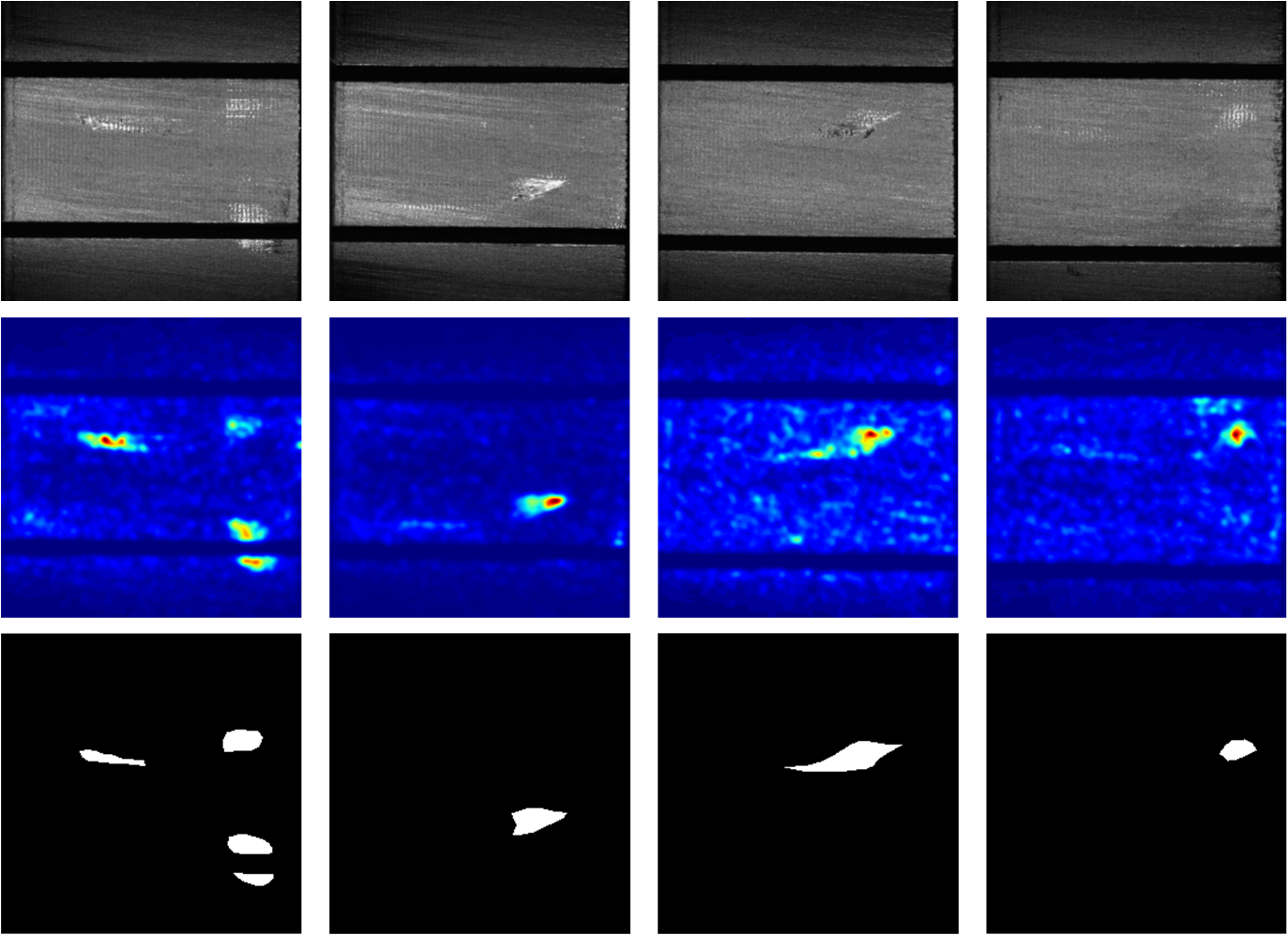}
    \caption{Examples of commutator surface defect segmentation results of proposed CMA-AE. From top to bottom are the input defective images, the error maps and the ground truth, respectively.}
    \label{fig:industrial-samples}
\end{figure}
\subsubsection{Influence of MSFR}
The purpose of MSFR is to accurately segment defects and suppress noise by leveraging feature residual at multi scales. To verify the effectiveness, we removed the entire MSFR from CMA-AE.

Column A and E of Fig. \ref{fig:ablation_analysis} show an example of the influence of MSFR. Instead of using MSFR, the model (column E) utilizes the pixel difference between input defective and its reconstruction, resulting in inaccurate defect segmentation and a lot of noise. By leveraging multi-scale feature residual, noise can be effectively suppressed and defect segmentation in more accurate. Table \ref{table:ablation_analysis} presents the quantitative experimental results. CMA-AE (column A) improves the AuROC by 0.71\% compared to the model without MSFR (column E).

This experiment confirms that our proposed MSFR is indeed effective. By exploiting multi-scale feature residuals, MSFR no longer focus on small gap between pixels, but on anomaly scores in regions of different sizes. This helps us achieve more accurate defect segmentation and suppress noise.

\subsection{Industrial Application}
To validate the potential of CMA-AE in the industrial field, as shown in Fig. \ref{fig:industrial-equipment}, it is implemented in our automated optical inspection equipment to inspect commutator surface defects online, which comprises the camera, light source, commutator, etc. A commutator dataset comprising 443 defect-free images and 66 defective images was collected, of which 336 nominal samples are leveraged for training and 97 nominal samples and 66 anomalous samples are utilized for testing.

Some examples of the defect segmentation results obtained using the CMA-AE method are shown in Fig. \ref{fig:industrial-samples}. The proposed CMA-AE is capable of inspecting and locating various defects accurately, which reveals its potential in the particular application.

\section{Conclusion}
In this paper, we proposed an unsupervised learning method CMA-AE for surface defect detection. This method is only trained on positive samples and artificial negative samples without any real negative samples. It is very important for the initial stage of the industrial production lines where negative samples are extremely scarce. 

Academically, we divide the key points of reconstruction-based anomaly detection methods into anomaly foreground repair ability and normal background reconstruction ability. A novel CMAM module and a new method of two-stage training strategy are proposed to obtain the two abilities. Furthermore, MSFR is proposed for more accurate and reasonable defect segmentation. We observed that MSFR outperforms the traditional method based on pixel gaps between original image and reconstructed image. Because MSFR uses encoders as feature extractors, it is suitable for all the auto-encoder-based or GAN-based methods. 

Extensive experimental results on several typical anomaly detection datasets show that our method CMA-AE achieves state-of-the-art detection accuracy. The performances may be further enhanced by improving GAAGA, such as edge of the weakening. In future research, we will focus on improving GAAGA to simulate more realistic defects.

\bibliographystyle{IEEEtran}
\bibliography{reference}

\begin{IEEEbiography}[{\includegraphics[width=1in,height=1.25in,clip,keepaspectratio]{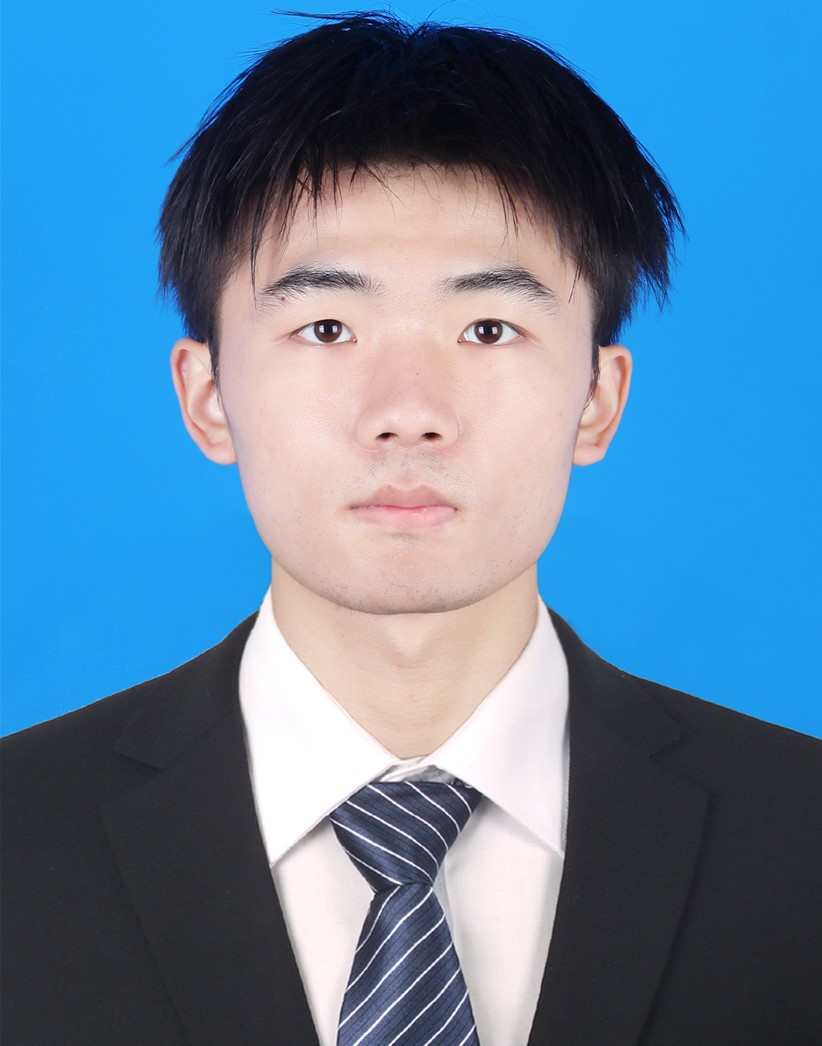}}]{Wei Luo} will receive a B.S. degree from the School of Mechanical Science and Engineering, Huazhong University of Science and Technology, Wuhan, China, in 2023. He is gong to pursue a Ph.D. degree with the Department of Precision Instrument, Tsinghua University. His research interests include deep learning, anomaly detection and machine vision.
\end{IEEEbiography}

\begin{IEEEbiography}[{\includegraphics[width=1in,height=1.25in,clip,keepaspectratio]{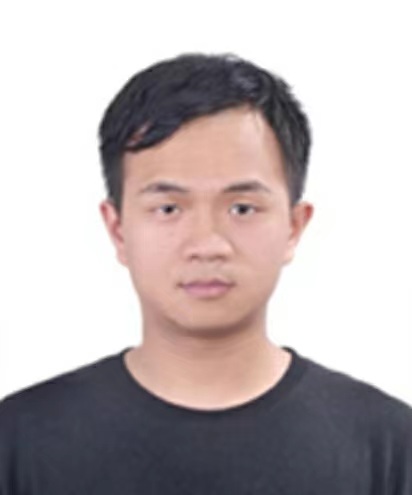}}]{Tongzhi Niu} received the B.S. degree in mechanical design, manufacturing, and automation from the Wuhan University of Technology, Wuhan, China, in 2018. He is currently pursuing the Ph.D. degree with the State Key Laboratory of Digital Manufacturing Equipment and Technology, Huazhong University of Science and Technology, Wuhan. His current research interests include intelligent manufacturing, defects detection, image processing, and deep learning.
\end{IEEEbiography}

\begin{IEEEbiography}[{\includegraphics[width=1in,height=1.25in,clip,keepaspectratio]{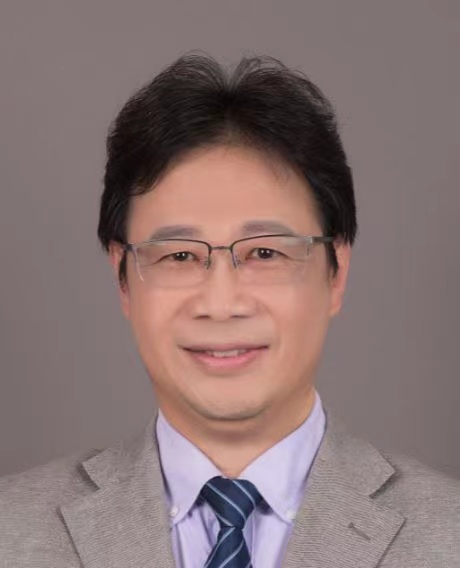}}]{Lixin Tang} received the B.S. and M.S. degrees from the Huazhong University of Science and Technology, Wuhan, China, in 1989 and 1992, respectively, and the Ph.D. degree from the University of Tsukuba, Tsukuba, Japan, in 2002. He is currently an Associate Professor with the School of Mechanical Science and Engineering, Huazhong University of Science and Technology. His current research interests include image process, computer vision, and pattern recognition.
\end{IEEEbiography}

\begin{IEEEbiography}[{\includegraphics[width=1in,height=1.25in,clip,keepaspectratio]{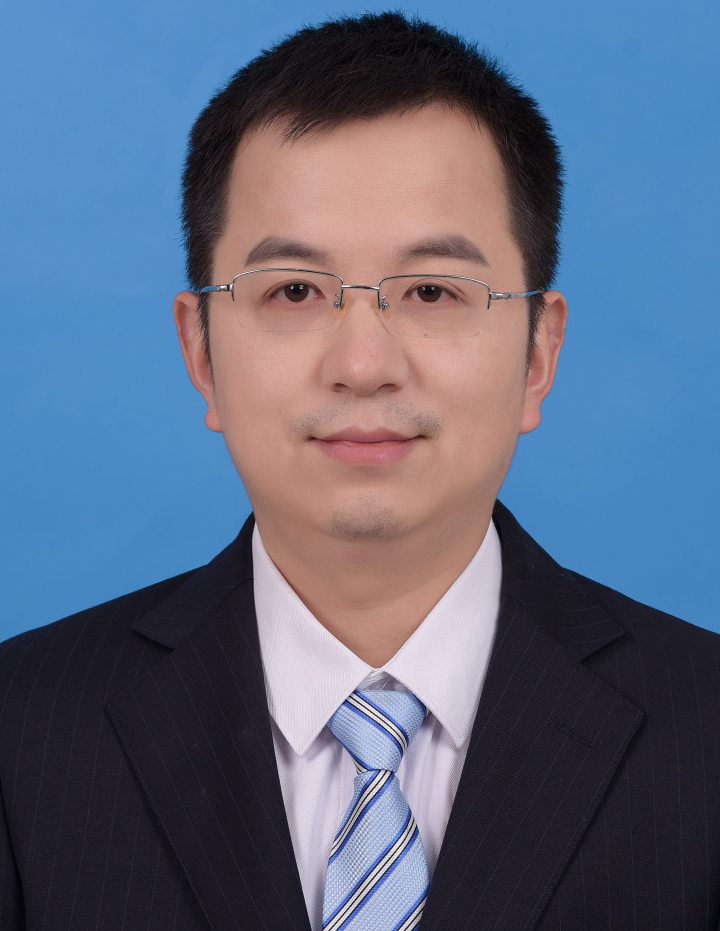}}]{Wenyong Yu} received an M.S. degree and a Ph.D. degree from Huazhong University of Science and Technology, Wuhan, China, in 1999 and 2004, respectively. He is currently a Associate Professor with the School of Mechanical Science and Engineering, Huazhong University of Science and Technology. His research interests include machine vision, intelligent control, and image processing.
\end{IEEEbiography}

\begin{IEEEbiography}[{\includegraphics[width=1in,height=1.25in,clip,keepaspectratio]{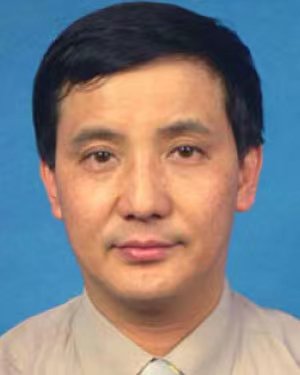}}]{Bin Li} received the B.S., M.S., and Ph.D. degrees in mechanical engineering from the Huazhong University of Science and Technology, Wuhan, China, in 1982, 1989, and 2006, respectively. He is currently a Professor with the School of Mechanical Science and Engineering, Huazhong University of Science and Technology. His current research interests include intelligent manufacturing and computer numerical control (CNC) machine tools.
\end{IEEEbiography}

\end{document}